%% file: main.tex
\documentclass[10pt,twocolumn,letterpaper]{article}

\usepackage{iccv}              


\definecolor{iccvblue}{rgb}{0.21,0.49,0.74}
\usepackage[pagebackref,breaklinks,colorlinks,allcolors=iccvblue]{hyperref}

\def\paperID{6532} 
\def\confName{ICCV}
\def\confYear{2025}

\title{Coarse Attribute Prediction with Task Agnostic Distillation for Real World Clothes Changing ReID}

\author{
\textbf{\url{https://ucf-crcv.github.io/ReID/}}\\\\
Priyank Pathak \hspace{20pt} Yogesh S Rawat\\
University of Central Florida\\
{\tt\small priyank@ucf.edu, yogesh@ucf.edu}
}

\usepackage{soul}
\usepackage{array}
 \usepackage{multirow} 
 \newcolumntype{P}[1]{>{\centering\arraybackslash}p{#1}}
 
\usepackage{makecell}

\newcommand{\B}[1]{{{\noindent\textbf{#1}}}}
\newcommand{\SB}[1]{{\noindent\underline{#1}}}

\usepackage{xcolor}
\usepackage{colortbl}
\definecolor{LGray}{gray}{0.9}
\sethlcolor{LGray}
\definecolor{ao(english)}{rgb}{0.0, 0.5, 0.0}
\newcommand{\Green}[1]{{\color{ao(english)}\textbf{#1}}}
\newcommand{\Red}[1]{{\color{red}\textbf{#1}}}

\newcommand{\IMPROV}[1]{{\color{ao(english)}\noindent\textsuperscript{\textbf{#1}}}}

\definecolor{mypink}{RGB}{255, 20, 147}
\hypersetup{
    colorlinks=true,
    linkcolor=blue,
    urlcolor=mypink,
}

\begin{document}
\maketitle
\input{sec/0_abstract}    
\input{sec/1_intro}

\input{sec/related}

\input{sec/method}

\input{sec/experiment}

\input{sec/result}

\input{sec/ablation}

\input{sec/conlcusion}

{
    \small
    \bibliographystyle{ieeenat_fullname}
    \bibliography{main}
}

\newpage

\maketitlesupplementary
\tableofcontents

\input{Supp/arch}

\input{Supp/motivation}

\input{Supp/training}

\input{Supp/vis}

\end{document}

%% file: sec/0_abstract.tex
\begin{abstract}
This work focuses on Clothes Changing Re-IDentification (CC-ReID) for the real world. 
Existing works perform well with high-quality (HQ) images, but struggle with low-quality (LQ) where we can have artifacts like pixelation, out-of-focus blur, and motion blur.
These artifacts introduce noise to not only external biometric attributes (e.g. pose, body shape, etc.) but also corrupt the model's internal feature representation. 
Models usually 
cluster LQ image features together, making it difficult to distinguish between them, leading to incorrect matches.
We propose a novel framework \textbf{Robustness against Low-Quality (RLQ)} to improve
CC-ReID model on real-world data.
RLQ relies on Coarse Attributes Prediction (CAP) and Task Agnostic Distillation (TAD) operating in alternate steps in a novel training mechanism. 
CAP enriches the model with \emph{external} fine-grained attributes via coarse \textit{predictions}, thereby reducing the effect of noisy inputs. 
On the other hand, TAD enhances the model’s \emph{internal} feature representation by bridging the gap between HQ and LQ features, via an 
external dataset through task-agnostic self-supervision and distillation.
RLQ outperforms the existing approaches by 1.6\%-2.9\% Top-1 on real-world datasets like LaST, and DeepChange, while showing consistent improvement of 5.3\%-6\% Top-1 on PRCC with competitive performance on LTCC.
\textit{\textbf{Github: \url{https://github.com/ppriyank/RLQ-CGAL-UBD}}}
\end{abstract}

%% file: sec/1_intro.tex
\section{Introduction}
\label{sec:intro}

Person Re-IDentification (ReID) involves identifying a person from a gallery of media, categorized as Short-Term and Long-Term (`Clothes Changing', `CC') ReID. The former identifies people with similar backgrounds/clothes while the latter identifies across locations, time, and clothing. Short-term ReID is relatively easier, with non-biometric cues (background and timestamps) helping in identification~\cite{wang2019spatial}.
 We focus on image-based CC-ReID counteracting real-world artifacts, without any non-biometric shortcuts.

\begin{figure*}[t]
\centering
\begin{subfigure}{.18\textwidth}
\centering
\includegraphics[height=1.9cm]{{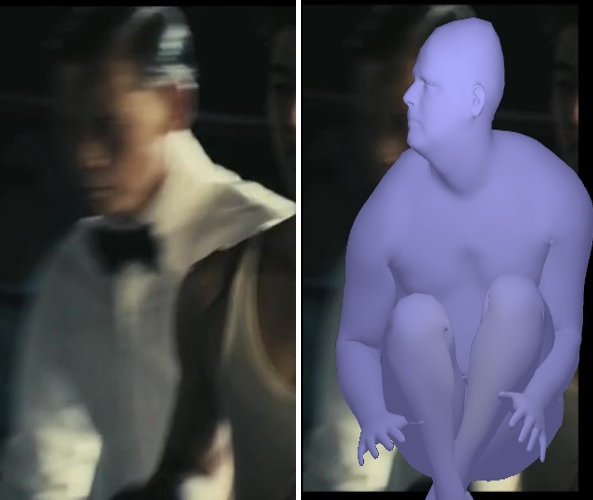}}
\caption{ \centering Motion Blur: Wrong 3D Body \& Pose (SMPL~\cite{10.1145/3596711.3596800}) }
\label{fig:body_fail}
\end{subfigure}
\hfill
\begin{subfigure}{.26\textwidth}
\centering
\includegraphics[height=1.9cm]{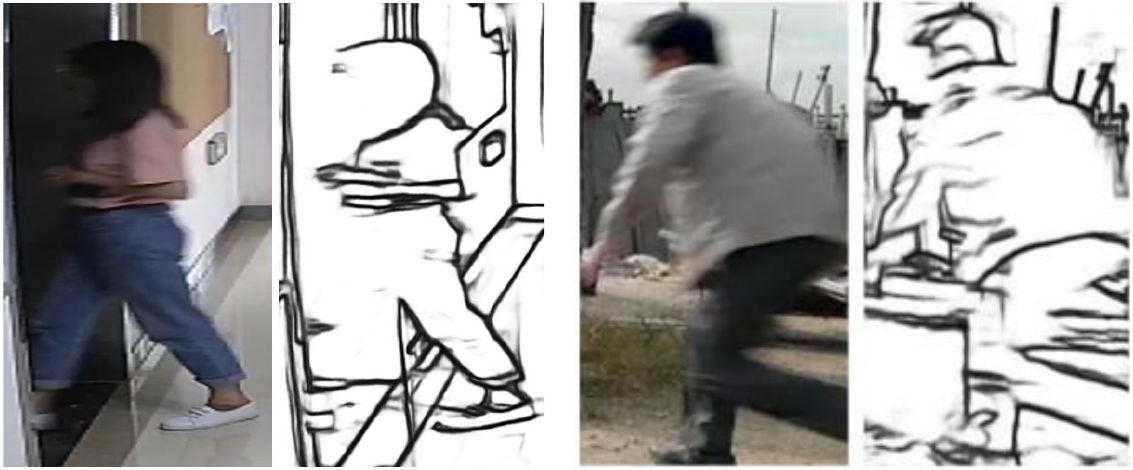}
\caption{ \centering Motion Blur:  Partial Contours (DexiNed~\cite{xsoria2020dexined}) }
\label{fig:contour_fail}
\end{subfigure}
\hfill
\begin{subfigure}{.18\textwidth}
\centering
\includegraphics[height=1.9cm]{{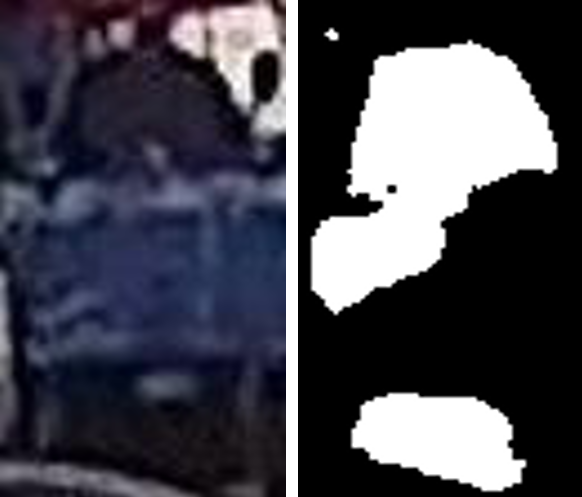}}
\caption{ \centering Pixelation: Noisy Silhouette (SCHP~\cite{li2020self}) }
\label{fig:sil_fail}
\end{subfigure}
\hfill
\begin{subfigure}{.18\textwidth}
\centering
\includegraphics[height=1.9cm]{{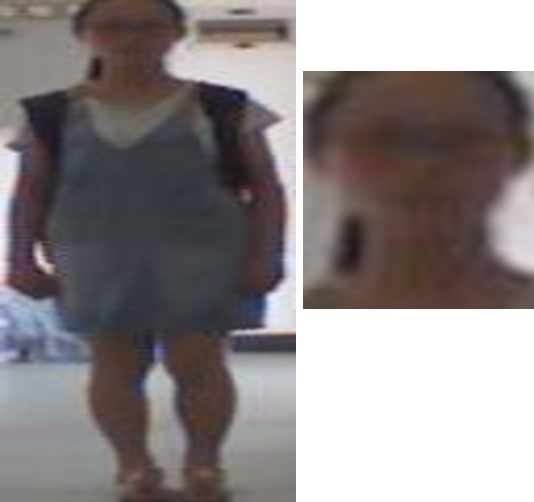}}
\caption{ \centering Out-of-focus: Blurry Face (RetineNet~\cite{Deng2020CVPR})}
\label{fig:face_fail}
\end{subfigure}
\hfill
\begin{subfigure}{.18\textwidth}
\centering
\includegraphics[height=1.9cm]{{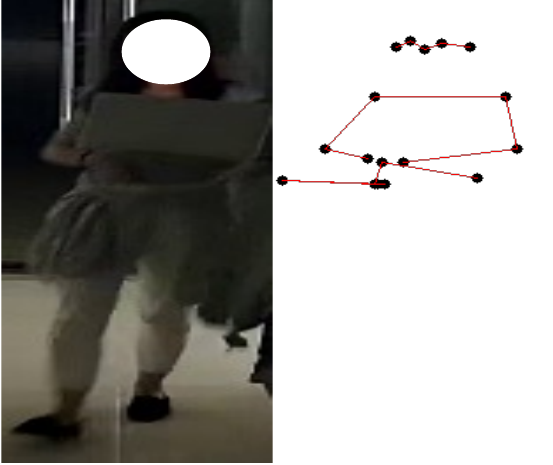}}
\caption{ \centering Pixelation: Partial Pose (AlphaPose~\cite{fang2017rmpe})}
\label{fig:pose2_fail}
\end{subfigure}
\caption{
\textbf{Noisy fine-grained attributes}: Samples from LaST, LTCC, DeepChange, PRCC dataset. 
}
\label{fig:Aux_Model_Fail}
\vspace{-7pt}
\end{figure*}

\begin{figure*}[t]
\centering
\begin{subfigure}{.58\textwidth}
\centering
\includegraphics[height=3.6cm]{{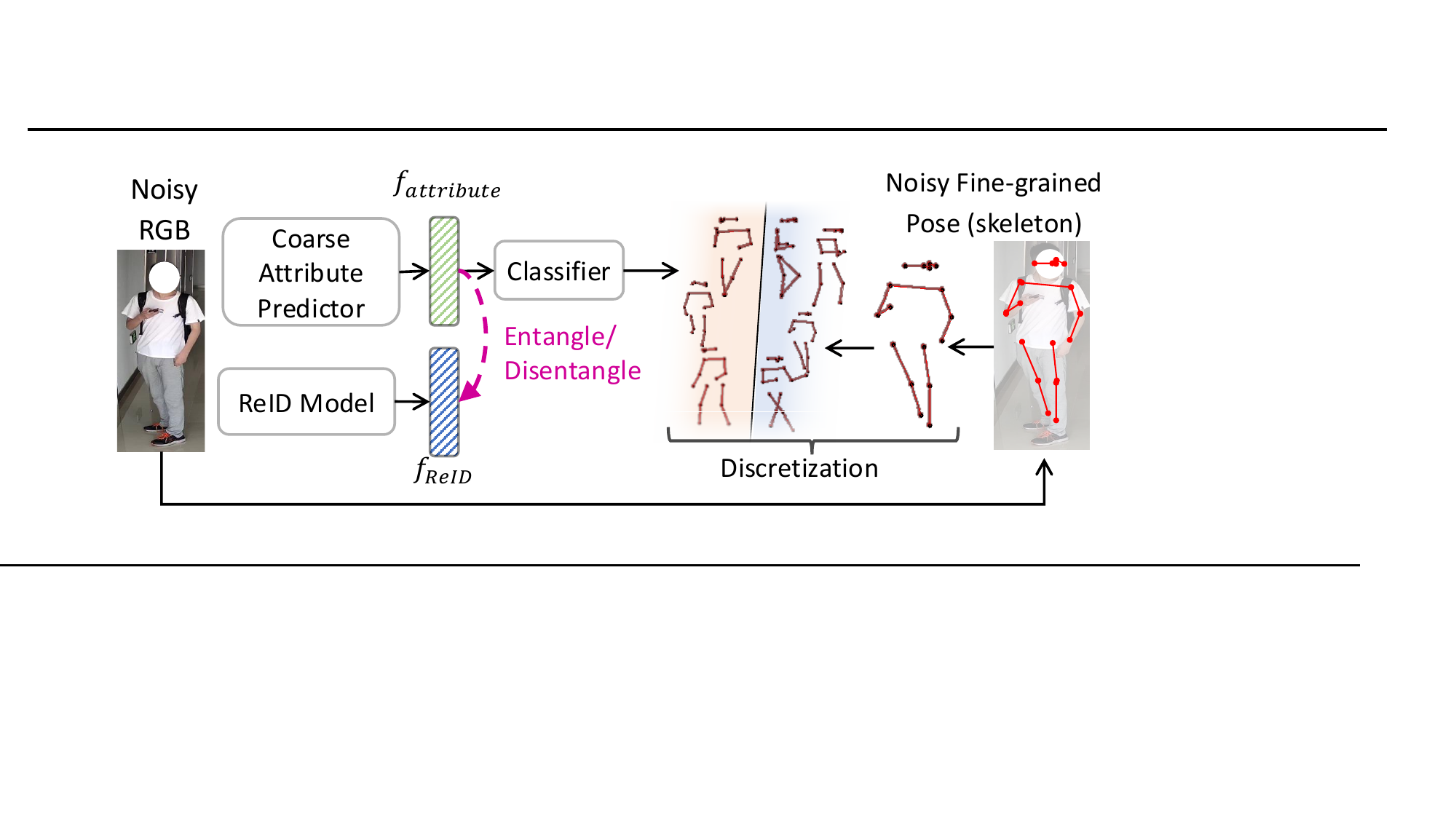}}
\caption{ \textbf{Coarse Attribute Prediction (CAP)} }
\label{fig:cgal_idea}
\end{subfigure}
\hfill
\begin{subfigure}{.36\textwidth}
\centering
\includegraphics[height=3.6cm]{{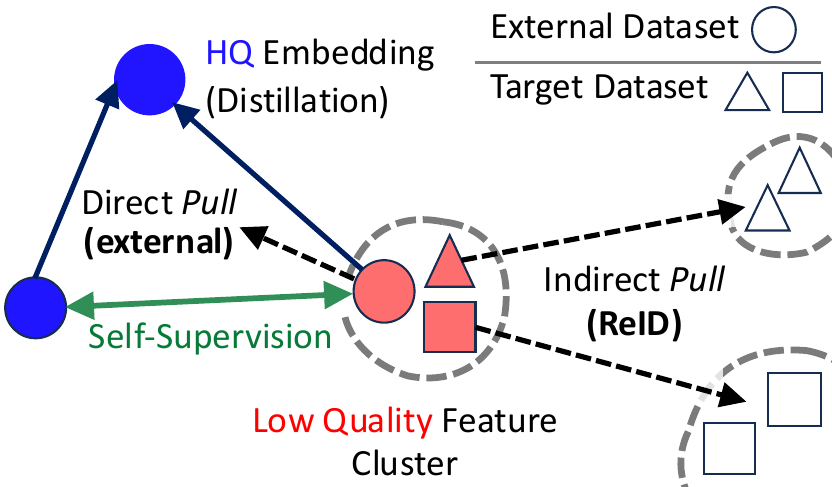}}
\caption{ \textbf{Task Agnostic Distillation (TAD)} }
\label{fig:ubd_idea}
\end{subfigure}
\caption{
\textbf{Robustness against Low-Quality (RLQ)} (a) Fine-grained attribute (\eg pose) is discretized/clustered into coarse classes which the model predicts using RGB images for learning attribute feature representation. Classification is an easy task that avoids input noise from attributes.
(b) External ($
{\bigcirc}$) HQ embeddings pull LQ ones towards it via self-supervision \& distillation, dissolving the LQ cluster on the target dataset ($\triangle$, $\square$), especially helpful when the target dataset doesn't have HQ-LQ pairs. 
}
\vspace{-8pt}
\end{figure*}

Existing approaches often rely on external fine-grained attributes (\eg silhouettes~\cite{Jin_2022_CVPR}, body shapes~\cite{Wang_2022_ACCV}, contours~\cite{9469545}, \etc) as auxiliary inputs. 
These attributes, produced by high-quality (HQ) pre-trained models, have not seen much success on low-quality (LQ) datasets. 
When applied to real-world LQ images, which include \emph{occlusions}, \emph{pixelation}, \emph{motion blur}, and \emph{out-of-focus (OOF)} blur, these attributes tend to be noisy (\cref{fig:Aux_Model_Fail}). 
Artifacts not only negatively impact auxiliary models but also ReID models themselves.
This results in a well-studied low-resolution issue, where models 
cluster LQ images regardless of identities, leading to indistinguishable erroneous matches. 

To address these issues we propose \textbf{Robustness against Low-Quality (RLQ)} framework which improves 
CC-ReID on real-world data.
RLQ \textit{alternates} between alleviating input noise from \emph{external} fine-grained attributes via \textbf{Coarse Attributes Prediction (CAP)}, and 
reducing the negative impacts of LQ images on the model's \emph{internal} features representation via \textbf{Task Agnostic Distillation (TAD)}. 
These two strategies when applied to our ReID model, enrich the model with fine-grained attributes (CAP) and distinguish LQ images (TAD).

CAP is proposed as an alternative to feeding additional input noise, via fine-grained attributes, while still learning their representation.  
Instead, feature representation for these \textit{external attributes} is learned by predicting these attributes as coarse discrete labels, directly from input (\cref{fig:cgal_idea}) \ie RGB-only inference. 
Here we focus on two attributes: gender and pose to show the effectiveness of CAP. 
We classify images into pose clusters, thereby disentangling pose (pose robustness), and gender classes, entangling gender awareness.

In addition to external attributes, ReID models themselves need robustness against real-world artifacts. 
Existing research~\cite{8695003, li2019recover} uses pairs of high-quality (HQ) and low-quality (LQ) images to alleviate the clustering of LQ features. 
However, separating HQ images from those with artifacts \eg pixelation in an upscaled image without HQ-LQ pairs is a challenging task. 
Our novel `TAD' (\cref{fig:ubd_idea}) bridges the domain gap between the LQ and HQ image feature, using distillation and self-supervision, via synthetic HQ-LQ 
image pairs generated from an external dataset. 
This pulls synthetic LQ features towards their HQ counterpart, implicitly helping dissolve the LQ feature clusters on our target dataset, thus helping the model distinguish LQ images better.

In summary, 
(1) We introduce a novel framework Robustness against Low-Quality (RLQ) for real-world CC-ReID. 
RLQ alternatively applies Coarse Attribute Prediction (CAP) and Task Agnostic Distillation (TAD) on the base ReID model for robustness against low-quality artifacts. 
(2) CAP entangles gender and disentangles pose, \emph{external} fine-grained attribute representation, without the additional input noise from these attributes.
(3) TAD enhances the model's \emph{internal} feature representation of low-quality images, via an external high-quality dataset reducing LQ image feature clusters.
(4) RLQ outperforms existing approaches on three popular CC-ReID benchmarks, namely PRCC, LaST, and Deepchange, with competitive performance on the LTCC. 
The extensive ablation study provides an analysis of the performance benefits of CAP and TAD.

%% file: sec/related.tex
\section{Related Work}
\label{sec:related}
\noindent \textbf{Clothes Changing Person (CC) ReID} 
Person ReID with same cloth~\cite{gao2018revisiting, Pathak2020FineGrainedR,9714137} is a well-researched topic.
Following the surge of CC-ReID~\cite{qian2020long,shu2021large, davila2023mevid}, we primarily focus on image-based CC-ReID. 
Our design adopts a multi-branch design~\cite{Hong_2021_CVPR, 9469545, 10.1007/978-981-99-7549-5_16} where one branch is dedicated to clothing changes, while the other focuses on biometric features. 

\noindent \textbf{Attributes:}
Multi-modal methods entangle fine-grained attributes (silhouettes~\cite{zhu2024sharc, Jin_2022_CVPR}, body shape~\cite{Wang_2022_ACCV}, body parts~\cite{Zhao_Liu_Chu_Lu_Yu_2021, ijcai2022p212}, contour~\cite{10.1007/978-3-030-87358-5_32}, skeleton~\cite{qian2020long, Hong_2021_CVPR}, face~\cite{arkushin2022reface, 9897243}, and 3D shape~\cite{10.1007/978-3-031-18907-4_3})
\emph{as input} via an external model to learn biometrics.
These attributes are essential in solving ReID, \eg pose robustness, but can also introduce additional input noise for real-world low-quality data.
Similar to generative models~\cite{liu2023learning, zheng2019joint, 9566823} learning representations by generating attributes (decouple (\eg pose, clothing) in a latent space), our approach predicts attributes, for learning representations. 
Predicting attributes reduces additional input noise, while coarse-grained attributes are easier to predict in the real world than generating fine-grained attributes. \vspace{2pt} 


\noindent \textbf{Low Resolution:} Prior research ~\cite{talreja2019attribute, 10042552} primarily addresses `pixelation' in face recognition.
However, the real world includes additional artifacts like blurring, 
which renders super-resolution~\cite{chen2019learning, JBRIMCRPReId} and traditional low resolution ineffective, with unreliable HQ-LQ image pairs.
Our solution is to create synthetic low-quality images~\cite {9098036, jiao2018deep} with real-world artifacts using an external HQ dataset, bridging the gap between the two domains.

\begin{figure*}[t]
\centering
\begin{subfigure}{.47\textwidth}
\centering
\includegraphics[height=2.15cm]{{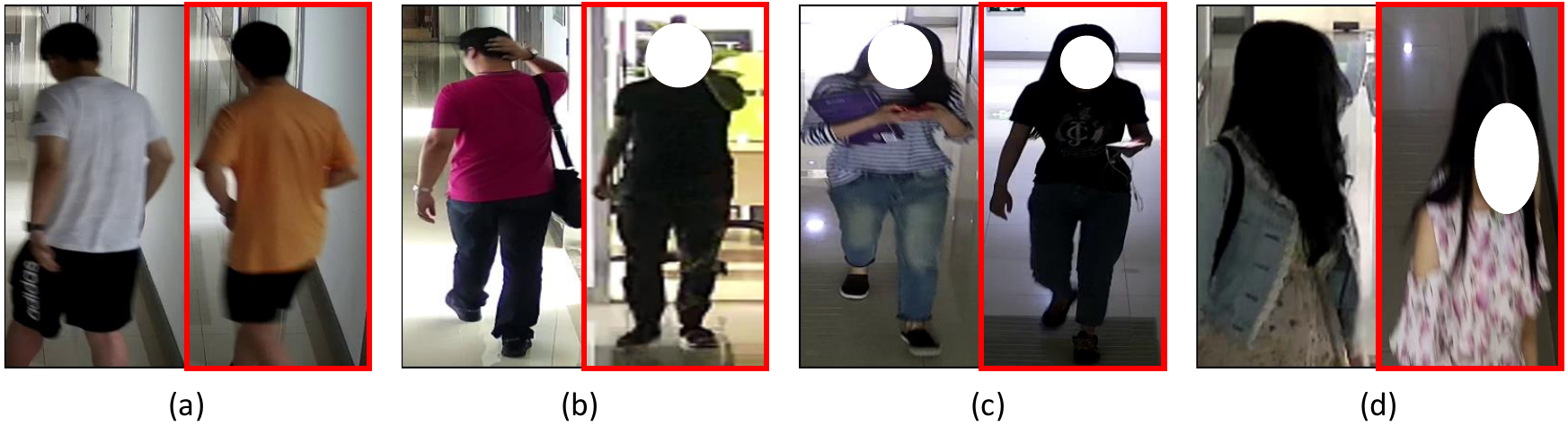}}
\caption{ Top-1 mismatch (shown in red) with similar Pose }
\label{fig:pose_mistake}
\end{subfigure}
\hfill
\begin{subfigure}{.47\textwidth}
\centering
\includegraphics[height=2.15cm]{{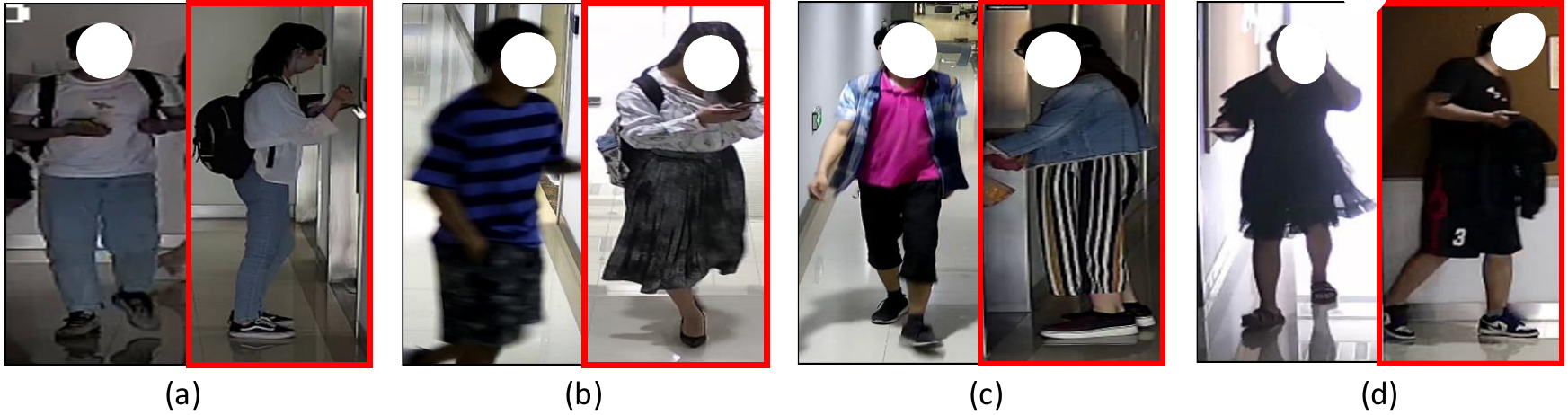}}
\caption{ Top-1 mismatch with opposite gender }
\label{fig:gender_mistake}
\end{subfigure}
\vspace{-5pt}
\caption{
\textbf{ReID against Pose and Gender:} LTCC mis-matches (in red) for CAL \cite{gu2022clothes}.
}
\label{fig:mistakes_in_CAL}
\vspace{-5pt}
\end{figure*}

%% file: sec/method.tex
 \begin{figure*}[!tb]
  \centering
  \includegraphics[width=0.95\linewidth]{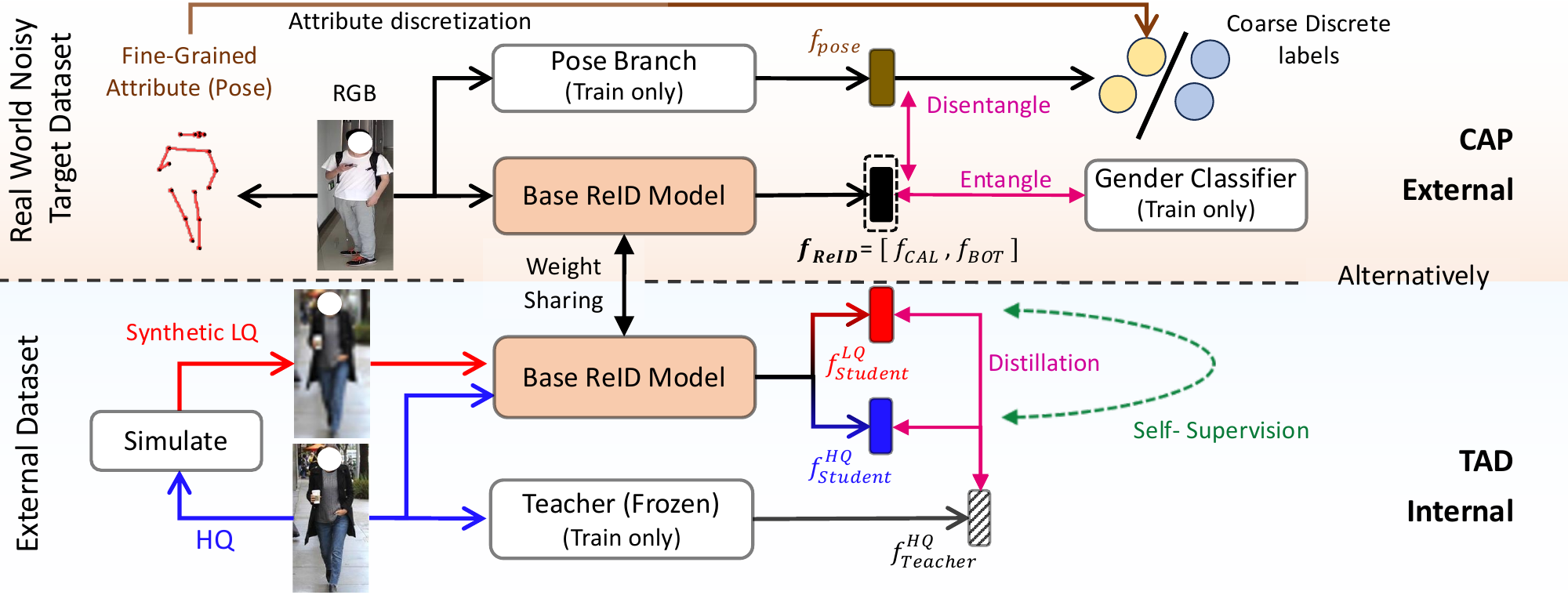}
  \caption{\textbf{Robustness against Low-Quality (RLQ)}: 
  Base ReID model is our baseline with RGB-only input (black arrows). 
  The base model is trained alternately between:  
  \textit{Upper}: Coarse Attributes Prediction (CAP) trained on target ReID dataset with train-only pose branch for pose disentanglement (pose clusters generated via AlphaPose (brown)) and gender classifier (gender assigned manually) for gender entanglement.  
  \textit{Bottom}: Task Agnostic Distillation (TAD) trains student Base model using teacher Base model (pink, distillation) and self-supervision on classification logits (green). 
  \textbf{Both CAP and TAD (White modules) are train-only, with Base Model $\boldsymbol{f_{ReID}}$ used during RGB-only inference}. The colored arrow indicates input. 
  }
  \label{fig:basemodel}
  \vspace{-5pt}
\end{figure*}

\section{Method}
\noindent \textbf{Overview:}
Robustness against Low-Quality (RLQ), focuses on real-world CC-ReID via Coarse Attributes Prediction (CAP) to enrich the model with \emph{external} fine-grained attributes, and Task Agnostic Distillation (TAD) for enhancing the model's \emph{internal} feature robustness on low-quality images (\cref{fig:basemodel}). 
During Training, the Base model (baseline) is trained using clothes-augmented RGB images, \textbf{with alternating steps of CAP} (pose, and gender) \textbf{and TAD} (external dataset, and teacher model). During Testing, \textbf{its RGB-only inference}, with \textit{no external attribute}. 
\textbf{Note:} Our focus in this work is not on removing
clothes bias; for that, we have the Base Model (CAL branch and cloth augmentations), instead, our contribution is to induce robustness against low-quality artifacts as present in Real World (inherently CC-ReID). 

\subsection{Proposed Baseline: Clothes Agnostic Base Model (BM)}
\label{sec:base_model}
We adopt one of the most popular open-source CC-ReID model CAL \cite{gu2022clothes} for learning clothes invariant features.
\textit{CAL focuses on clothing} where a clothes classifier predicts clothing labels, while an adversarial loss penalizes correct clothes predictions, thereby removing clothes-relevant information.
However, CAL lacks an understanding of pose and gender (\cref{fig:mistakes_in_CAL}). 
Coupled with its \emph{sensitive} nature, it is not trivial to integrate additional losses in this architecture.
Hence we adopt a two-branch architecture  ~\cite{Hong_2021_CVPR, 10036012}, where Bag-of-tricks (BOT~\cite{luo2019bag}) branch is added to CAL (ResNet-50 backbone split identically after the second block).
BOT Branch trains with additional losses and transfers knowledge to the CAL branch via a shared backbone.
Concatenation of features from both branches is used during inference $\boldsymbol{f_{ReID}}$ (=$[f_{BOT}, f_{CAL}]$). 
Train-only Identity (ID) classifiers predict ID labels on each branch and KL-divergence on these ID logits 
($\mathcal{L}_{ID,KL}$) helps transfer knowledge across the branches indirectly.
We use \textit{clothes augmented images}~\cite{9956160, 9802835} via body parsing~\cite{li2020self} \textit{to help the vanilla BOT branch with clothes changing}.
Collectively these two branches constitute our \textbf{Base Model} which serves as our baseline. Detailed description is provided in \textit{Supplmentary}. \Cref{tab:trip_aug} indicates that this additional clothes augmentation and two branch structures offer no performance boost, but simply stabilities CAL for additional training modules.  
Basemodel (CAL and BOT branch are used during inference).

\begin{table*}[!t]
\renewcommand{\arraystretch}{1}
\setlength\tabcolsep{3pt}
  \centering
  \scalebox{0.89}{
  \begin{tabular}{l|
c| cc| cc| cc| c|c|c|c|c}
    \toprule
    \multicolumn{2}{c|}{\multirow{2}{*}{Dataset}} & \multicolumn{2}{c|}{\# of IDs} &
    \multicolumn{2}{c|}{\# of Imgs} & \multicolumn{2}{c|}{\# of Clothes} & \multirow{2}{*}{Px.} & \multirow{2}{*}{OOF} & \multirow{2}{*}{\makecell{M. \\ Blur} } & \multirow{2}{*}{ \makecell{Brief \\Description}} & \multirow{2}{*}{ \makecell{Evaluation \\Protocol}}
    \\ \cline{3-8}
    \multicolumn{2}{c|}{} & Train & Test & Train & Test & Train & Test & & & & & \\
    \midrule
    \multirow{3}{*}{\rotatebox[]{90}{LQ}} 
    & Celeb-ReID\cite{huang2019celebrities} (TAD-only)$\dagger$ & 632 & 420 & 20,208 & 13,978 & 20,208 & - &  & & & HQ celebrity images & - \\ 
    & LTCC\cite{qian2020long} & 77 & 75 & 9,576 & 7,543 & 256 & 221 & \checkmark & \checkmark & \checkmark & 12 indoor views & CC, General \\
    & PRCC\cite{yang2019person} & 150 & 71 & 17,896 & 10,800 & 300 & 142 & \checkmark & \checkmark & & 3 indoor views & CC, SC \\
    \hline 
    \multirow{2}{*}{\rotatebox[]{90}{VLQ}} 
    & DeepChange\cite{xu2021deepchange} & 450 & 521 & 75,083 & 80,483 & 1,219 & 1,336 & \checkmark & \checkmark  &  & 12 months, 17 views & General \\ 
    & LaST\cite{shu2021large} & 5,000 & 5,807 & 71,248 & 135,529 & 16,035 & -& \checkmark & \checkmark & \checkmark & web/movie images & General \\ 
    \hline    
  \bottomrule
  \end{tabular}
  }
  \vspace{-5pt}
  \caption{\textbf{Dataset, artifact \& Evaluation protocol}: \emph{Px.}: denotes pixelation, \emph{OOF}: denotes out-of-focus blur, \emph{M. Blur}: denotes motion blur, \emph{CC}: means clothes changing, \emph{SC}: means same clothes.
$\dagger$: \textbf{Celeb-reID} is chosen as the external dataset for TAD, (not for RLQ evaluation). 
PRCC and LTCC are real-world LQ datasets, while DeepChange and LaST are Very LQ datasets (VLQ). \textbf{All artifacts \emph{empirically observed}}.}
\label{tab:dataset}
\vspace{-8pt}
\end{table*}

\subsection{Coarse Attribute Prediction (CAP): External Attributes Features}
\label{sec:cgal}
The key motivation behind Coarse Attribute Prediction (CAP) is to learn \textit{essential external} fine-grained attributes without the additional input noise due to low-quality images.
Generative method ~\cite{liu2023learning} can learn attributes via predictions as well (\eg generate poses) for learning pose, however, classification is an easier task than generating real-world data.
Thus, predicting these attributes is an obvious choice that CAP achieves by classifying attributes as coarse discrete labels from raw RGB images as shown in \Cref{fig:basemodel} (upper). 
This not only reduces the input noise but also removes the test time dependency on external attributes. 
CAP can disentangle attributes (remove pose bias) or entangle them (gender awareness) into the Base Model (BM). 
CAP applies pose branch and gender classifier on the BOT branch of the BM during the training. \vspace{1pt}

\begin{figure}[t]
\centering
\begin{subfigure}{0.38\linewidth}
\centering
\includegraphics[width=\linewidth]{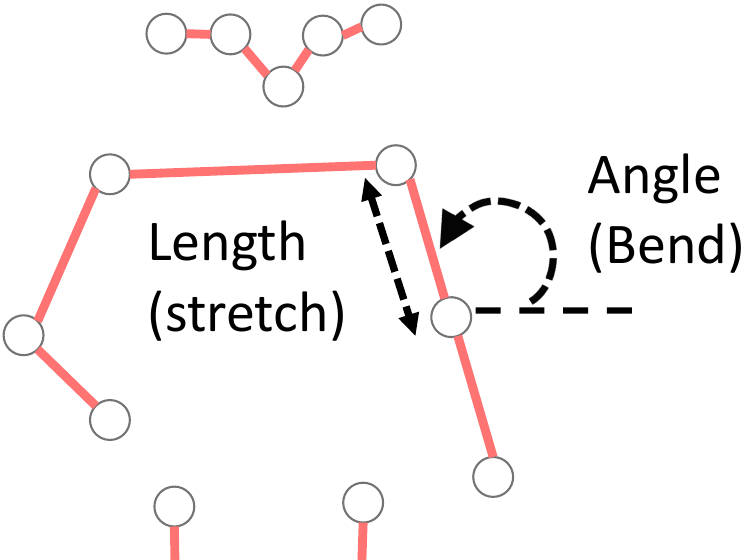}
\label{fig:pose_vector}
\end{subfigure}
\hfill
\begin{subfigure}{0.6\linewidth}
\centering
\includegraphics[width=\linewidth]{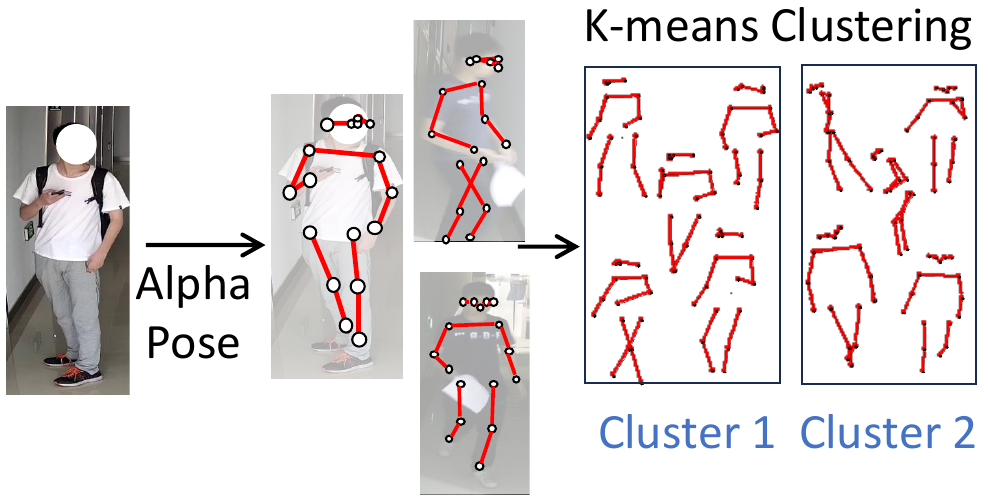}
\end{subfigure}
\vspace{-8pt}
\caption{\textbf{Pose Clustering}:
\textit{left}: Pose Vector (AlphaPose~\cite{fang2017rmpe}) consists of the length and angle of 13 body lines, which are used in clustering (K-means). \textit{right}: 2 pose clusters shown. 
}
\label{fig:pose_clustering}
\vspace{-8pt}
\end{figure}

\noindent \textbf{Pose Branch:}
\label{sec:posemodule}
Pose embedding $f_{POSE}$, is learned by classifying RGB images into pose clusters (similar to how ImageNet classification helps learn image class embeddings).
Unlike pose \emph{entanglement}~\cite{Nguyen_2024_WACV}, 
we believe pose is temporary and shouldn't play any role in identifying an individual, and thus disentangled.
By maximizing $f_{POSE}$ angular distance with $f_{BOT}$ towards orthogonality \ie minimize absolute cosine similarity, ($\mathcal{L}_{Cos}$), we disentangle pose bias from the features~\cite{kim2023feature}.
\begin{equation}
    \mathcal{L}_{Cos} = \big| \text{ Cosine-Similarity}( f_{BOT}, f_{POSE} ) \hspace{0.07cm} \big|
    \label{eq:pose}
\end{equation}
To further ease disentanglement, the pose branch shares the backbone with the Base Model (split after the first block) keeping their features in the same feature space. 

AlphaPose~\cite{fang2017rmpe} is used to generate fine-grained 2D skeletons.
Contrary to the traditional clustering~\cite{Hong_2021_CVPR, Nguyen_2024_CVPR} that \textbf{represent pose} via the \textbf{location of body joints}, we additionally include line joining joints referred to as body lines, which consist of \textbf{stretch} (length of body lines), and \textbf{bend} (angle of body lines) as shown in~\cref{fig:pose_clustering}.
These add a more fine-grained nuance to the pose vector. 
Pose vectors are then clustered using L2 distance-based K-means, assigning each image a discrete pose cluster. 
\textit{Supplementary} visualizes pose clusters. The pose branch then clusters raw RGB images into these clusters learning the pose representation.
\vspace{8pt}

\noindent \textbf{Gender Classifier}
\label{sec:gendermodule}
Gender is often used as soft-biometrics~\cite{galiyawala2018person} for filtering out wrong genders.
Gender Classifier on $f_{BOT}$
classifies RGB images into binary gender labels \ie male vs female, entangling gender-based cues (\eg male/female body shape) into ReID.

\subsection{Task Agnostic Distillation (TAD): Model's Internal Features}
\label{sec:distillationstep}
Models cluster LQ image features as erroneous look-a-like.
Traditional solutions of low-quality features imitating high-quality ones are not applicable without clean separable HQ-LQ pairs.
Addressing these, TAD leverages external HQ datasets to bridge the gap between the two domains, via synthetic HQ-LQ pairs. 
As shown in \cref{fig:ubd_idea}, external dataset HQ features pull LQ features via self-supervision and distillation. 
This implicitly helps ReID pull LQ feature on the target dataset, dispersing low-quality clusters. 
This enhances models \emph{internal} robustness against LQ artifacts (helping `see better'). 
Unlike previous work~\cite {jiao2018deep, 9098036}, TAD is \textbf{1) trained parallelly to the ReID task, in alternate epochs}, while 
\textbf{2) the external dataset plays no role in supervising ReID}, aka ``task-agnostic''.
\vspace{8pt}

\begin{table*}[!th]
  \centering
  \renewcommand{\arraystretch}{0.9}
\setlength\tabcolsep{3pt}
\scalebox{0.95}{
\begin{tabular}{ 
c|l|c| cc| cc| cc| cc}
\toprule
\multicolumn{2}{c|}{\multirow{3}{*}{Method}} & 
\multirow{3}{*}{ \makecell{Additional \\Attributes}}
& \multicolumn{4}{c|}{LTCC (\%)} 
& \multicolumn{4}{c}{PRCC (\%)} \\
\cline{4-11}
\multicolumn{2}{c|}{ } &  & 
\multicolumn{2}{c|}{CC} & \multicolumn{2}{c|}{General} &
\multicolumn{2}{c|}{CC} & \multicolumn{2}{c}{SC}\\
\cline{4-11}
\multicolumn{2}{c|}{ } & &R-1$\uparrow$ &mAP$\uparrow$  &R-1$\uparrow$ & mAP$\uparrow$ &R-1$\uparrow$ &mAP$\uparrow$ &R-1$\uparrow$ &mAP$\uparrow$ \\
    \hline 
\multirow{5}{*}{\rotatebox[]{90}{Prev.  Work}}   
& BSGA (BMVC'22) \cite{bmvc_lcccpr} & S &  - & - & - & - & \SB{61.8} & 58.7 & 99.6 & 97.3 \\ 
&3DInv$_{\text{CAL}}$ (ICCV'\citeyear{liu2023learning}) & P & 40.9 & 18.9 & - &  - & 56.5 & 57.2 & - & -  \\ 
&CCFA (CVPR'23)~\cite{han2023clothing} & - & 45.3 & 22.1 &  75.8 & \SB{42.5} &  61.2 & 58.4 & 99.6 & 98.7 \\ 
& CVSL (WACV'\citeyear{Nguyen_2024_WACV}) & P &  44.5 & 21.3 & 76.4 &  41.9 & 57.5 & 56.9 & 97.5 & 99.1 \\ 
& CCPG (CVPR'24)~\cite{Nguyen_2024_CVPR} & P & \SB{46.2} & \B{22.9} & \B{77.2} & \B{42.9} & \SB{61.8} & 58.3 & \B{100} & 99.6\\ 
& CLIP3D$\ddag$ (CVPR'24)~\cite{Liu_2024_CVPR} & B,Text & 42.1 &  \SB{21.7} & - & - &  60.6 &  59.3 &- & - \\
\hline 
\multicolumn{2}{c|}{ \textit{Baseline:}  CAL (CVPR'22)~\cite{gu2022clothes}} & - & 40.1 & 18.0 & 74.2 & 40.8 & 55.2 & 55.8 & \B{100} & \SB{99.8} \\
\multicolumn{2}{l|}{\textit{Baseline:} Base Model}  & S & 
41.1 & 20.3 & 75.7 & 41.4 & 
58.0 & 59.4 & \B{100} & \SB{99.8} \\ 
\hline  
\rowcolor{LGray}
 & CAP  & S, P, G & 43.4\IMPROV{+2.3} & 20.7\IMPROV{+0.4}  & 74.7\noindent\textsuperscript{-1.0} & 
41.9\IMPROV{+0.5} & 60.2\IMPROV{+2.2} & 60.3\IMPROV{+0.9} & \B{100} & \B{99.9}\\
\rowcolor{LGray}
 &TAD  & S & 43.4\IMPROV{+2.3} & 21.6\IMPROV{+1.3} & 76.1\IMPROV{+0.4} & 41.8\IMPROV{+0.4} & 61.2\IMPROV{+3.2} & \SB{61.1}\IMPROV{+1.7} & \B{100} & \B{99.9}\\ 
\rowcolor{LGray}
\multirow{-3}{*}{\rotatebox[]{90}{Our}}  & 
RLQ (CAP + TAD) & S, P, G & \B{46.4}\IMPROV{+5.3} & $21.5$\IMPROV{+1.2} & $\SB{76.9}$\IMPROV{+1.2} & 41.8\IMPROV{+0.4} & $\B{64.0}$\IMPROV{+6.0} & $\B{63.2}$\IMPROV{+3.8} & \B{100} & \SB{99.8}\\ 
\hline 
  \bottomrule
  \end{tabular}
}
\vspace{-5pt}
   \caption{\textbf{Comparison on LTCC and PRCC (\%)}. 
  ``-'' means results were not reported. $\uparrow$ means higher the number better the performance.
  Shorthand notations: `S': silhouettes/body parsing, `P': Pose (2D/3D), `B': Body Shape (2D/3D), and 'G': Gender.
CLIP3D$\ddag$~\cite{Liu_2024_CVPR} is short for `CLIP3DReID'.
`\IMPROV{+x}' shows x\% increment over baseline base model.
}
\label{tab:ltcc_prcc_results}
\vspace{-5pt}
\end{table*}

\begin{table*}[!th]
  \centering
\begin{minipage}{.5\linewidth}
\centering
  \renewcommand{\arraystretch}{0.9}
\setlength\tabcolsep{3pt}
\scalebox{0.92}{
\begin{tabular}{l|c|cc}
\toprule
Method & 
Attributes & R-1$\uparrow$  & mAP $\uparrow$ \\
\hline 
M2Net-F (MM'22)~\cite{liu2022long} & S, C & \SB{57.2} & 19.6 \\
IMS+GEP
(TMM'23)~\cite{10237321} & - & 55.1 &  18.3 \\ 

MCSC-CAL (TIP'24)~\cite{10472423} & - & 56.9 & \SB{21.5} \\ 
CLIP3D (CVPR'24)~\cite{Liu_2024_CVPR} &B,Text  &  56.7 &  20.8\\ 
 \hline 
 \textit{Baseline:} CAL (CVPR'22)~\cite{gu2022clothes}
 & - & 54.0 &  19.0 \\ 
 \textit{Baseline:} Base Model
 & S & 57.7 & 21.4 \\ 
\hline 
\rowcolor{LGray}
RLQ (Our)
& S, P, G  &
$\B{58.8}$ & $\B{22.1}$ \\
\hline 
\bottomrule
\end{tabular} 
}
\vspace{-5pt}
\caption{\textbf{DeepChange General Results (\%)} (\Cref{tab:ltcc_prcc_results} Formatting). 
}
\label{tab:DeepChange}
\end{minipage}%
\hfill
\begin{minipage}{.5\linewidth}
\centering
  \renewcommand{\arraystretch}{1}
\setlength\tabcolsep{3pt}
\scalebox{0.92}{
\begin{tabular}{l|c|cc}
\toprule
Method & Attributes & R-1$\uparrow$ & mAP$\uparrow$ \\
\hline 
LaST
(TCSVT'22)~\cite{shu2021large}) & - & 71.0 & 28.0 \\ 
MCL
(MM'22)~\cite{10.1145/3503161.3547900} & - & \SB{75.0} & 22.7 \\ 
\citeauthor{10.4018/IJWSR.322021}
(AIMS'23)~\cite{10.4018/IJWSR.322021} & - & 68.9 & 24.1 \\ 
IMS+GEP (TMM'23)~\cite{10237321} & - & 73.2 & \SB{29.8}\\ 
\hline 
\textit{Baseline:} CAL (CVPR'22)\cite{gu2022clothes}
 & - & 73.7 &  28.8 \\   
 \textit{Baseline:} Base Model
 & S & 75.7 & 31.7 \\
\hline
\rowcolor{LGray}
RLQ (Our) & S, P, G & \B{77.9} & \B{35.3} \\
\hline
\bottomrule
\end{tabular}
}
\vspace{-5pt}
\caption{\textbf{LaST General Results (\%)} (\Cref{tab:ltcc_prcc_results} Formatting).}
\label{tab:LaST}
\end{minipage}     
\vspace{-10pt}
\end{table*}

Matching HQ-LQ features, risks pulling HQ embeddings towards LQ clusters~\cite{9137263}, thus an \emph{indirect} softer approach is adopted. 
Student HQ \& LQ features are pulled toward the teacher's frozen HQ embeddings, reinforced by self-supervision on classifiers logits \ie HQ logit match LQ logits, instead of features.
Teacher's \textbf{{distillation}} eliminates the need for labels from the external dataset, making any HQ dataset suitable regardless of the target ReID task (task-agnostic).
Additionally, the teacher's fixed anchor embedding helps maintain the separability of features, which serves two purposes: 
(1) \textbf{Prevents Collapsing}: Collapsing external dataset embeddings into a lump.
(2) \textbf{No multi-tasking}: allows a small subset to fix LQ clusters without full training on the external dataset. 
Empirical evidence in ablation. 
We hypothesize that the absence of task-specific awareness on the target dataset (\ie ``task agnostic'' (no ReID)) combined with the \textbf{self-supervision} pull being as strong as distillation pull, helps mitigate distribution shift from the external dataset. \vspace{8pt}

\Cref{fig:basemodel} (bottom) shows TAD. 
External HQ dataset pre-trains teacher base model and generates synthetic HQ-LQ pairs: $(x_{HQ}, x_{LQ})$. We use Celeb-ReID~\cite{huang2019celebrities} without labels as an external HQ dataset. The \emph{frozen} teacher only sees $x_{HQ}$ producing normalized CAL and BOT features 
$[ T^{HQ}_{CAL}, T^{HQ}_{BOT}]$, while the student sees both $(x_{HQ}, x_{LQ})$ generating normalized features as $[S^{HQ}_{CAL},S^{HQ}_{BOT}]$, and 
$[S^{LQ}_{CAL},S^{LQ}_{BOT}]$. 
Classifiers produce ID logits $[Y^{\delta}_{CAL, ID}, Y^{\delta}_{BOT,ID}]$ and CAL's clothing logits $[Y^{\delta}_{CAL,CL}]$, where $\delta \in \{HQ, LQ\}$. MSE loss (pink in the figure) \textit{distills} features, while KL divergence (green in the figure) is used for soft \textit{self-supervision}
\begin{align}
\mathcal{L}_{D}&=\hspace{-0.3cm}\sum_\delta^{\{HQ,LQ\}}\hspace{-0.2cm}\| T^{HQ}_{CAL}-S^{\delta}_{CAL}\|_2^2 + 
\| T^{HQ}_{BOT}-S^{\delta}_{BOT}\|_2^2\label{eq:mse}\\
\mathcal{L}_{S}&=\hspace{-0.1cm}KL\bigl(
Y^{HQ}_{CAL,CL} \bigl\| Y^{LQ}_{CAL,CL} \bigl) + 
\hspace{-0.6cm}
\sum_\alpha^{\{CAL,BOT\}} \hspace{-0.4cm} KL \bigl(
Y^{HQ}_{\alpha,ID} \bigl\| Y^{LQ}_{\alpha,ID} \bigl)
\label{eq:kl}  
\end{align}

\subsection{Overall Learning Objective}
Baseline inherits the traditional losses of BOT and CAL branches. These include Label-smoothed cross-entropy loss (CE) for both branches identity classifiers (identity classification), 
on CAL branch clothes classifier CE and CAL loss (clothes prediction and adversarial loss), and Triplet loss for $f_{BOT}$.
We \textbf{introduce} KL-divergence loss ($\mathcal{L}_{ID,KL}$) between the ID logits of both branches.
\textit{CAP} introduces CE loss for pose ($\mathcal{L}_{POSE,CE}$) and gender ($\mathcal{L}_{GEN,CE}$) classification, and $\mathcal{L}_{Cos}$, (\cref{eq:pose}) for pose disentanglement.
\textit{TAD} losses are distillation $\mathcal{L}_{D}$, (\cref{eq:mse}), and self-supervision $\mathcal{L}_{S}$ (\cref{eq:kl}).
The weight for each loss is 1.

%% file: sec/experiment.tex
\section{Experiments and Results}
\noindent \textbf{Dataset \& Evaluation}
\Cref{tab:dataset} describes the four image CC-ReID datasets. Datasets are evaluated using Top-1 (\emph{R-1}, CMC) and mAP evaluation. 
The evaluation protocol: (i) \textit{Cloth-Changing (CC)}:  Query-Gallery wears different clothes. (ii) \textit{Same-Clothes (SC)}: Query-Gallery have matching clothes. (iii) \textit{General:} All query and gallery samples. \vspace{8pt}

\begin{table*}[!th]
\centering
\begin{minipage}{.33\linewidth}
\centering
\renewcommand{\arraystretch}{1.1}
\setlength\tabcolsep{3pt}
\scalebox{0.95}{
\begin{tabular}{@{ }P{0.9cm}@{}| @{ }P{1cm}@{}| cc|cc}
\toprule
\multirow{2}{=}{Cloth Aug} & \multirow{2}{=}{Triplet Loss} & \multicolumn{2}{c|}{CAL} & \multicolumn{2}{c}{BM} \\ 
\cline{3-6}
& & R-1 & mAP & R-1 & mAP \\
\hline 
& & \textbf{40.1} & \textbf{18.0} & 36.7 & 16.0 \\ 
\hline 
\checkmark & & 36.9 & 18.0 &  40.1 & 17.9\\
\hline 
 & \checkmark  & 34.7 & 16.6 & 38.8  & 18.3  \\ 
 \hline 
\checkmark & \checkmark  & 35.5 & 17.0 & \textbf{41.1}\cellcolor{LGray}  & \cellcolor{LGray}\textbf{20.3}  \\
\hline 
\bottomrule
\end{tabular}
}
\caption{\textbf{CAL Sensitivity I:} Triplet Loss \& Clothes Augmentation (Aug.) impact on CAL and BM.}
\label{tab:trip_aug}
\end{minipage}%
\hfill
\begin{minipage}{0.29\linewidth}
\centering
\renewcommand{\arraystretch}{1.1}
\setlength\tabcolsep{3pt}
\begin{tabular}
{c|cc|cc}
\toprule
\multirow{2}{*}{Branch}
&\multicolumn{2}{c|}{BM$+$G}&\multicolumn{2}{c}{BM$+$P} \\ 
\cline{2-5} 
& R-1 & mAP & R-1 & mAP \\    
\hline 
\cellcolor{LGray} BOT & \cellcolor{LGray} \textbf{43.5} & \cellcolor{LGray} \textbf{21.7} & \cellcolor{LGray} \textbf{42.5} & \cellcolor{LGray} \textbf{20.2} \\ 	
CAL & 43.4 & 20.8 & \textbf{42.4} & \textbf{20.2 } \\	
Both & 41.9	& 20.4 & 41.5 & 19.5 \\
\hline 
\bottomrule
\end{tabular}
\caption{
\textbf{CAL Sensitivity II:}
Position of Gender (G) entangle \& Pose (P) disentangle for CAL, BOT \& both branches.}
\label{tab:gender_pose_pos}
\end{minipage}%
\hfill 
\begin{minipage}{0.33\linewidth}
\centering
\renewcommand{\arraystretch}{1}
\setlength\tabcolsep{3pt}
  \begin{tabular}{c|c|  cc}
    \toprule
   TAD  Variants & ED & R-1 & mAP \\
    \hline 
 BM (Baseline) & - & 41.1 & 20.3 \\ 
 \hline 
BM + LQ Aug. & & 41.2 & 19.5 \\ 
$SS_{MSE}$ + TAD  & \checkmark & 43.3 & 21.2 \\ 
$SS_{MSE}$ [NT] & \checkmark & 32.4 & 16.3 \\ 
LTCC+$SS_{MSE}$ [NT] & & 38.3 & 18.5 \\ 
\rowcolor{LGray}
TAD (Reported) & \checkmark &   \textbf{43.4} & \textbf{21.6} \\   
  \hline 
\bottomrule
\end{tabular}
\caption{\textbf{TAD Variants}:
  ED: external dataset, [NT]: No Teacher.
  $SS_{MSE}$: Self-supervision on features (MSE).
  Explained in Ablation. 
  }
\label{tab:ubd_varaint}
\end{minipage}%
\vspace{-6pt}
\end{table*}

\begin{table*}[!th]
\begin{minipage}{.22\linewidth}
\centering
\renewcommand{\arraystretch}{1.1}
\setlength\tabcolsep{3pt}
\begin{tabular}{c|cc}
\toprule
\multirow{2}{*}{\makecell{TAD \\ HQ Datasets}} & \multicolumn{2}{c}{PRCC} \\ 
\cline{2-3}
& R-1 & mAP  \\
\hline 
BM  &	58.0 & 59.4\\
NTU &	61.1	& 59.8\\
\rowcolor{LGray}
Celeb-ReID & \textbf{61.2} & \textbf{61.1}\\
\hline 
\bottomrule
\end{tabular}
\caption{
\textbf{HQ datasets in TAD: } 
NTU~\cite{liu2020ntu} dataset instead of Celeb-ReID}
\label{tab:ntu_tad}
\end{minipage}%
\hspace{2pt}
\begin{minipage}{.25\linewidth}
\includegraphics[height=3.3cm]{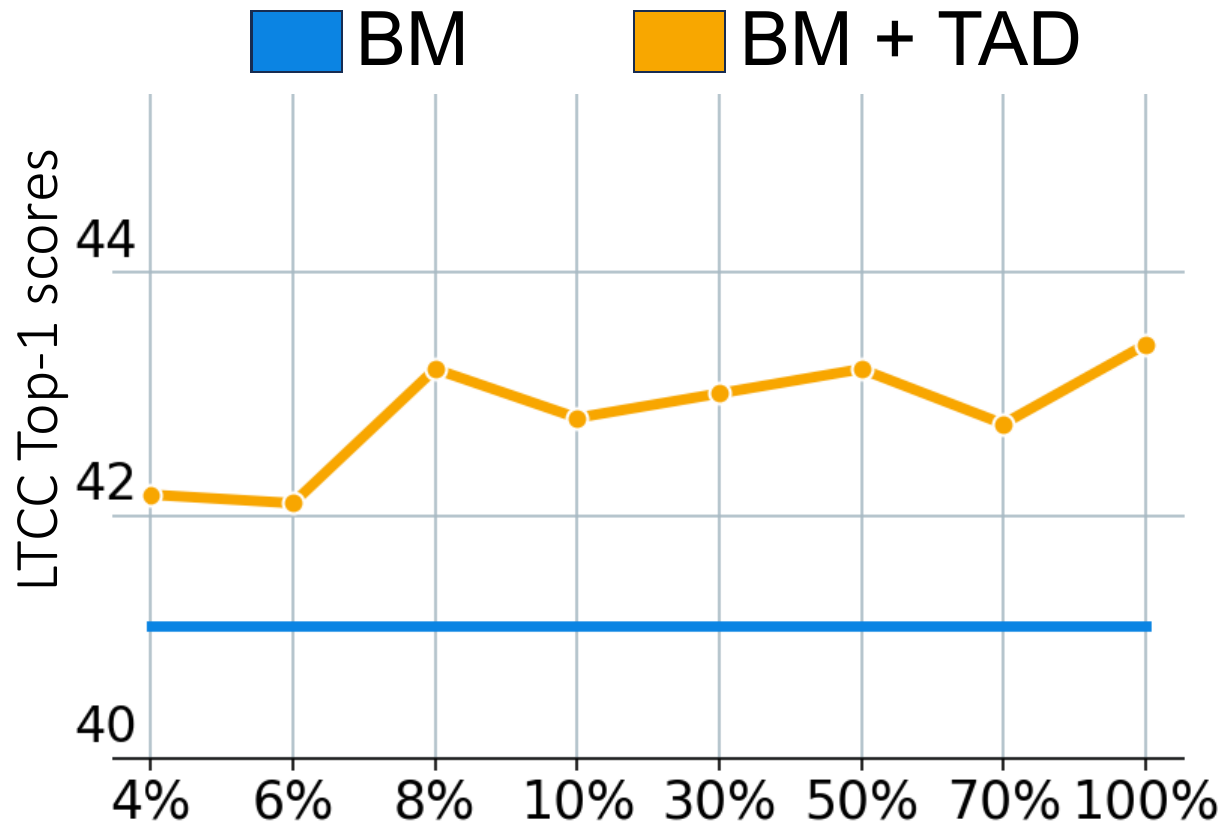}
\captionof{figure}{Celeb-ReID Sampling}
\label{fig:celeb_sampling}
\end{minipage}%
\hfill
\begin{minipage}{.48\linewidth}
\centering
\renewcommand{\arraystretch}{1.1}
\setlength\tabcolsep{3pt}
\scalebox{0.9}{
\begin{tabular}{c| c| cc| cc|| cc| cc}
\toprule
\multirow{3}{*}{Aux.} & \multirow{3}{*}{\#Par} & \multicolumn{4}{c||}{BM} & \multicolumn{4}{c}{BM + TAD} \\ 
    \cline{3-10}
    & & \multicolumn{2}{c|}{LTCC} & \multicolumn{2}{c||}{PRCC}  
    & \multicolumn{2}{c|}{LTCC} & \multicolumn{2}{c}{PRCC} \\
    \cline{3-10} 
    & & R-1 & mAP & R-1 & mAP & R-1 & mAP & R-1 & mAP \\
    \hline 
    BM & 45.9 & 41.1 & 20.3 & 58.0 & 59.4 & 43.4 & 21.6 & 61.2	& 61.1 \\ 
    +G & 45.9 & 43.5	& 21.7 & 58.8 & 59.8 & 44.8 & \textbf{22.0} & 62.4 & 62.6 \\ 
    		
    +P & 69.3 & 42.5 & 20.2 & 59.4 & 60.6  & 44.7 & 21.4 & 61.8 & 62.3 \\ 
    \rowcolor{LGray}
    +CAP & 69.3 & 43.4 & 20.7 & 60.2 & 60.3 & \textbf{46.4} & 21.5 & \textbf{64.0}& \textbf{63.2} \\ 
    \hline 
  \bottomrule
  \end{tabular}
  }
  \caption{\textbf{Contribution of architectural component}. G.: Gender, P: Pose, CAP: Gender + Pose, \#Par: parameters count in millions.}
\label{tab:arch_comp}
\end{minipage}%
\vspace{-10pt}
\end{table*}

\noindent \textbf{Implementation}
\label{sec:implementdetails}
Branches are ImageNet initialized (no pre-training), with output `Max-Avg' pooled  (concat of max and avg pool ~\cite{huang2021clothing}).
Model is trained end-to-end for 200 epochs (300 epochs for LaST) with 40 batch sizes. 
Gender is manually annotated as 1(male), and 0(female).
Pose vectors are clustered with an empirically found cluster size of 15 ($0^{th}$ cluster for noisy poses).
In PRCC and DeepChange, \textbf{pixelation} and \textbf{out-of-focus} (OOF) is simulated with $\frac{1}{2}$ probability. \textbf{Motion blur} is added for LTCC and LaST. 
Pixelation is downsampling to resolution $\in [16\times16, 64\times64]$, and resizing back.
\emph{OOF} blurring is Gaussian blur 
 with a random kernel $\in [5,21]$. 
\emph{Motion blur} is obtained by applying a random size kernel $\in [8,20] $ of 1s rotated randomly $\in [0^{\circ},180^{\circ}]$. 
More details are provided in \textit{Supplementary}.

%% file: sec/result.tex
\subsection{Results}
\label{sec:results}
Ensuring fair comparison with all methods, accuracy-boosting techniques that can be universally applied to all models (including ours),
\eg Re-Ranking~\cite{ZHANG2023109070}, Gallery enrichment~\cite{arkushin2022reface}, or Vivo learning~\cite{10364687} are not used in comparison.
All results follow the color scheme: \hl{Our results}, {{\textbf{SOTA}}}, and {{\underline{second-best}}}. 
Results of the Celeb ReID dataset can be found under Limitations.

\noindent \textbf{LTCC \& PRCC (Real World LQ)}:  \Cref{tab:ltcc_prcc_results} shows the comparison of our model with existing methods. 
Auxiliary attributes-based models in general outperform RGB-only models (CAL, CCFA).
We surpass the previous best on the PRCC dataset with competitive results on the LTCC dataset, giving consistent improvement on CC protocol over the base model as \underline{6.0\% Top-1 and 3.8\% mAP} on PRCC and \underline{5.3\% Top-1 and 1.2\% mAP} on LTCC.

\noindent \textbf{DeepChange, and LaST (Very LQ):} 
\Cref{tab:DeepChange} \& \Cref{tab:LaST} shows results on the 
DeepChange and LaST. 
There are only a few \emph{fine-grained attributes work on these two datasets}, partly because of their LQ artifacts training data. 
Attributes-related works include Contour (M2Net-F) and Text features w/ 3D body shape (CLIP3D), rest are heuristic-based works.
 RLQ outperforms the previous best by \underline{1.6\% and 2.9\%} Top-1, and \underline{0.6\% and 5.5\%} mAP on DeepChange and LaST datasets, respectively.
\textbf{Avoiding input noise / TAD enables us to use pose / gender like fine-grained attributes for datasets like DeepChange and LaST (among the only work to do so)}

%% file: sec/ablation.tex
\section{Ablations and Analysis}
\label{sec:ablation}

Following has, 
\emph{BM$+$G}: Base Model (BM) w/ only gender classifier. Similarly \emph{BM$+$P}: w/ only pose branch, \emph{BM$+$TAD}: w/ only TAD, and 
\emph{BM$+$CAP}: RLQ w/o TAD \ie BM$+$Pose$+$Gender.
All ablations are performed with CC protocol for the LTCC dataset (unless mentioned). LTCC is relatively smaller than LaST and DeepChange, making ablation easier. 
Design choices are \hl{highlighted}
\vspace{8pt}
\begin{figure*}
\centering
  \centering
  \includegraphics[width=\linewidth]{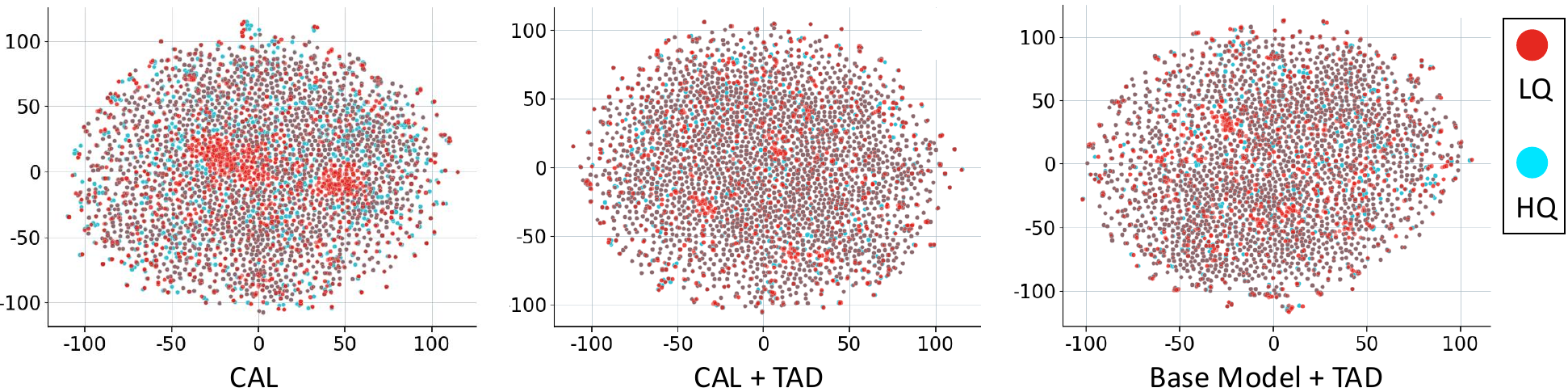}
  \captionof{figure}{\textbf{Visual evidence of robustness against LQ clusters}: t-SNE of pairs of HQ (blue) \& LQ (red) features for CelebReID dataset with synthetic artifacts. Lumps of Red dots (and not blue dots) indicate model cluster LQ features regardless of identities, with TAD reducing these clusters.  
  }
  \label{fig:lr_hr_cluster}
\vspace{-5pt}
\end{figure*}

\begin{figure*}[!tb]
\centering
\subfloat
{\includegraphics[height=4.2cm]{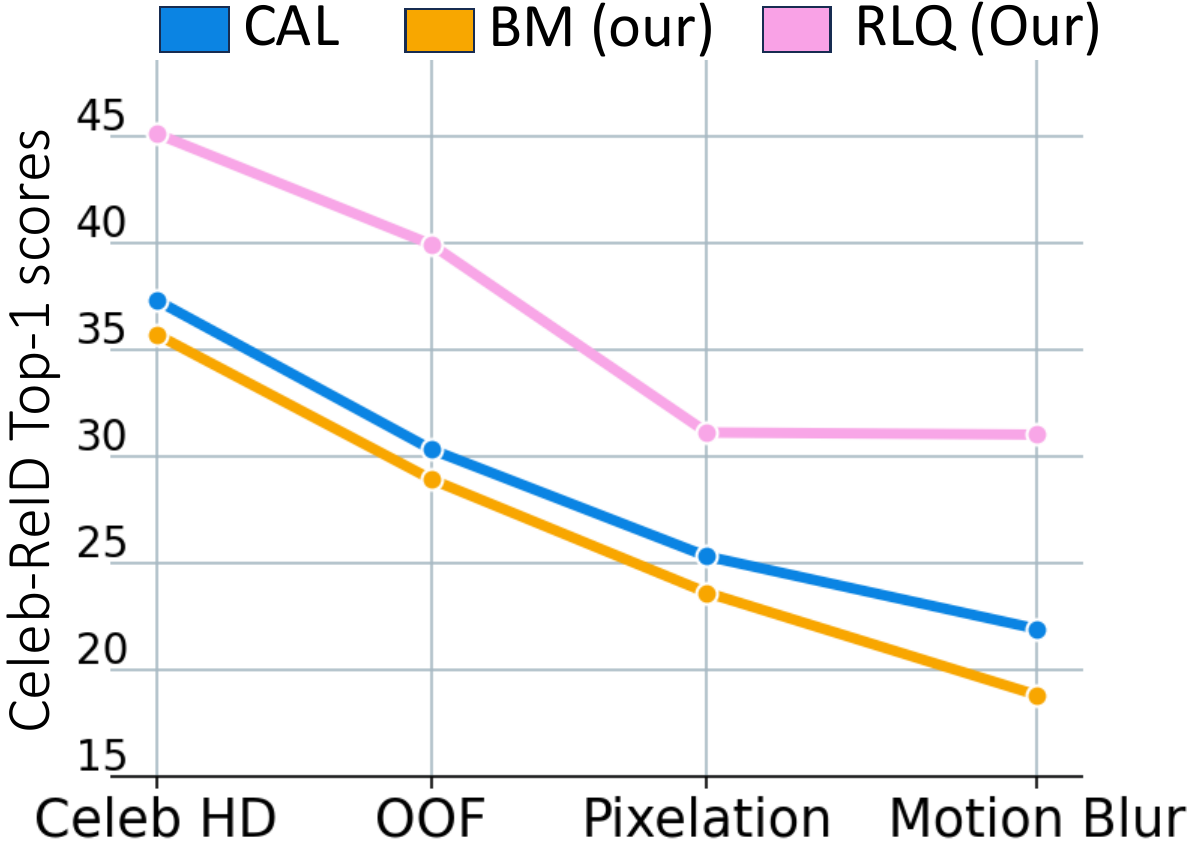}
}
\hfill
\subfloat
{\includegraphics[height=4.2cm]{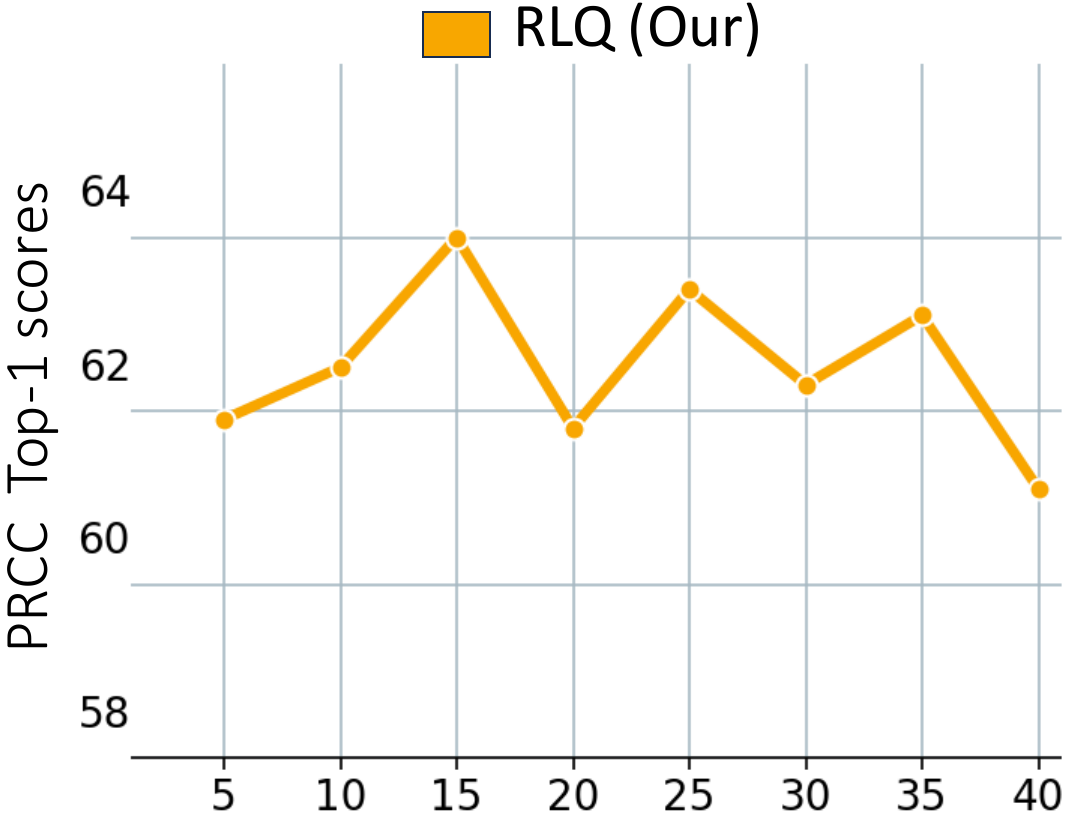}
}
\hfill
\subfloat{\includegraphics[height=4.2cm]{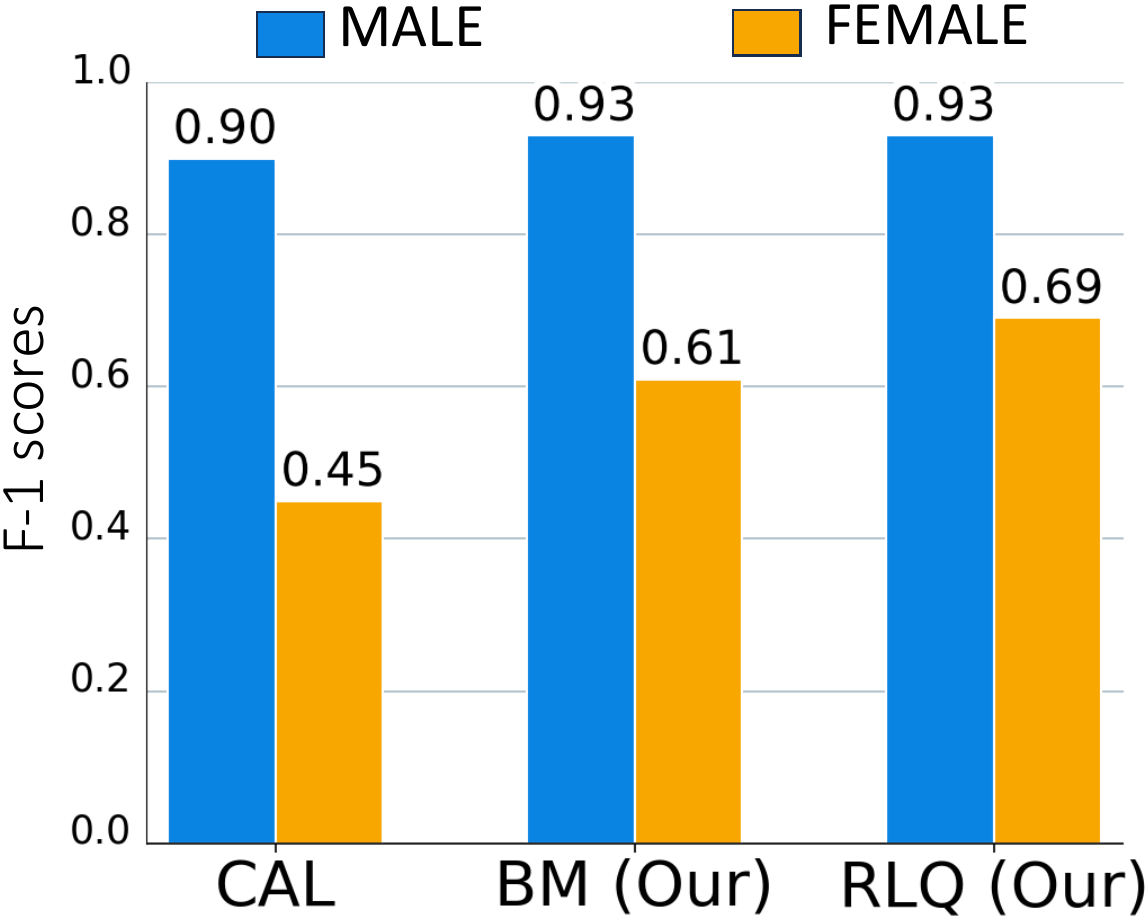}
}
\caption{
\textit{Left:} Impact of artifacts. Motion blur affects the performance most severely. RLQ (w/ UBD) outperforms CAL and Base model.
\textit{Mid:} Optimal K-means cluster of pose vectors for PRCC, at 15 clusters. 
\textit{Right:} F1-scores on gender classification, with RLQ being more gender aware in ReID than CAL and Base Model.}
\label{fig:cal_gender_error2}
\label{fig:prcc_pose_clusters}
\label{fig:cal_gender_error}
\label{fig:model_analysis}
\vspace{-10pt}
\end{figure*}

\noindent \textbf{Base Model (Cal Sensitivity)}
\label{sec:arch_choice}
\underline{\Cref{tab:trip_aug}} illustrates the \textit{incompatibility of CAL's adversarial loss} with triplet loss and clothes augmentations. This is further evident from \underline{\Cref{tab:gender_pose_pos}}, where pose disentanglement and gender entanglement perform optimally on the BOT branch. 
Base Model with \textbf{extra BOT branch and foreground augmentation does not have any performance boost} (Base Model 41.1\% vs vanilla RGB CAL 40.1\%) but simply \textbf{fixes the CAL's sensitivity problem}.
Design choices in \textit{Supplementary}. \vspace{8pt} \\
\noindent \textbf{TAD (Teacher \& External Dataset)}
\underline{\Cref{tab:ubd_varaint}} shows variations of TAD for improving robustness against LQ. 
Base Model with LQ augmentations (BM+LQ Aug.) doesn't impact the performance. 
Directly matching student's HQ-LQ features via MSE loss ($SS_{MSE}$), 
with teacher $SS_{MSE}$+TAD and without teacher $SS_{MSE}$[NT], offers no performance gain over TAD, while later dropping performance.
LTCC+$SS_{MSE}$[NT], $SS_{MSE}$ with noisy HQ-LQ pairs from LTCC instead of clean Celeb-ReID, has low performance. 
This indicates the need for clean HQ-LQ image pairs, via external dataset. 
Poor performance without the teacher {[NT]} indicates a possible feature collapse on the external datasets because of distribution shift, justifying the teacher's soft supervision for feature separability.
\vspace{6pt}

\noindent \textbf{TAD (Task-agnostic)}
\underline{\Cref{tab:ntu_tad}} shows the use of NTU~\cite{liu2020ntu}, another HQ dataset, in TAD instead of Celeb-ReID, as a source of synthetic HQ-LQ image pairs. NTU improvement is slightly inferior to Celeb-ReID, possibly because of limited viewpoints. 
\underline{\Cref{fig:celeb_sampling}} shows random sampling at $\sim$8\% of Celeb-ReID external dataset maintains consistent performance gain on TAD. This shows that the small amount of data with the teacher's soft supervision is effective in learning the HQ-LQ relationship, without the need for extensive multi-tasking on the external dataset. 
These indicate the ReID/task-agnostic nature of the external dataset.
\vspace{6pt}

\begin{figure*}[!tb]
\centering
\subfloat[\centering RLQ correctly identifies \textbf{Gender}]
{{\includegraphics[height=5.3cm]{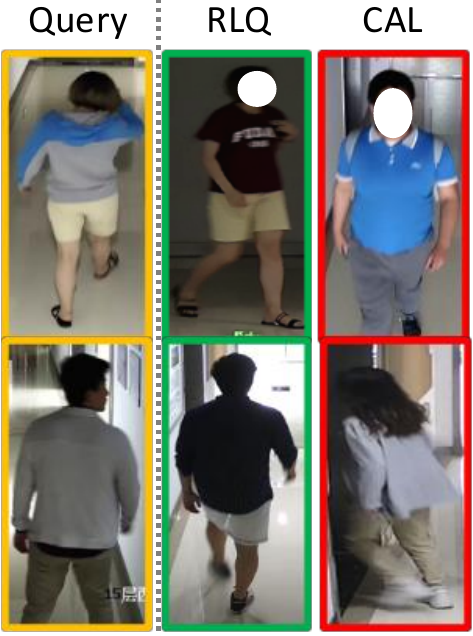}
\label{fig:cal_gender_error}
}}
\hfill
\subfloat[\centering RLQ is \textbf{Pose Robust}]{{\includegraphics[height=5.3cm]{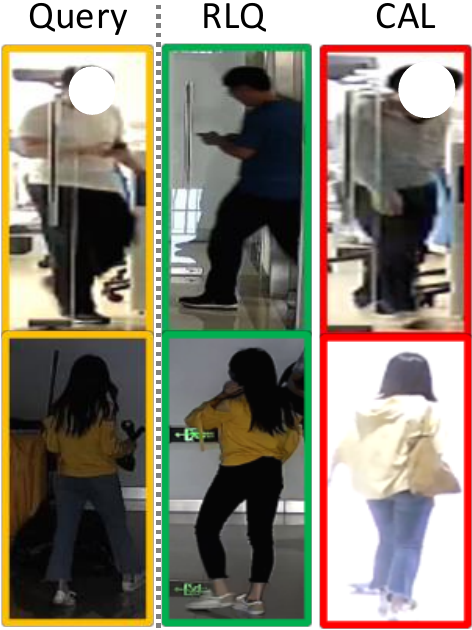}
\label{fig:cal_pose_error}
}}
\hfill
\subfloat[\centering RLQ is \textbf{Low Quality Robust}]{{\includegraphics[height=5.3cm]{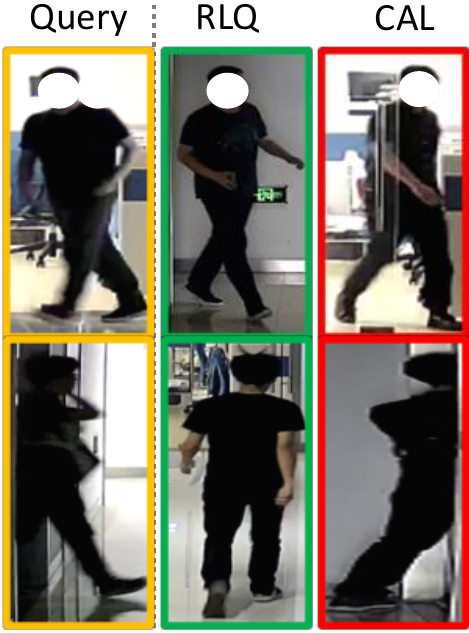}
\label{fig:cal_lr_error}
}}
\hfill
\subfloat[\centering \textbf{Both Wrong: }Gender \& Pose]{{\includegraphics[height=5.3cm]{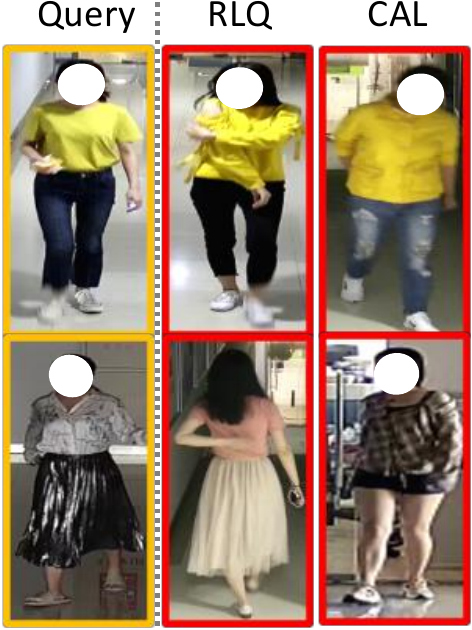}
\label{fig:both_error}
}}
\vspace{-5pt}
\caption{\textbf{RLQ vs CAL Visual Error Analysis:} 
LTCC Top-1 match shown, with \Green{correct} and \Red{incorrect} matches. General improvements where CAL makes a mistake while RLQ doesn't are: (a) Gender recognition (b) Pose robustness (c) Low-quality robustness (LQ images match with other LQ images because of LQ feature clustering). When both models make mistake (d), RLQ still correctly identifies gender and avoids similar pose.}
\label{fig:cal_vs_rql}
\vspace{-10pt}
\end{figure*}

\noindent \textbf{Each Component \& Number of Params}
\underline{\Cref{tab:arch_comp}} shows the performance of each component on the Base model, w/ and w/o TAD. Gender has the largest impact on the LTCC, while pose performs well on the PRCC. TAD gives a performance boost to each component. 
The addition of pose with gender (CAP) for LTCC, slightly reduces the performance. This may be because of noisy pose clusters, which the model can fix (see better) upon application of TAD.    
The final combination RLQ achieves the best performance. 

The gender classifier has minimal parameter overhead, while the pose branch results in a 50\% parameter gain.
A better baseline, unlike the sensitive CAL, would eliminate the need for the BOT branch, significantly reducing the parameters.
Also, 69M parameters are still relatively small compared to transformers-based approaches (LTCC sota) with ViT-Base/Swin-Base, with $\sim$87M parameters. 
Time analysis is tricky, as inference time is typically reported. TAD and external attributes are only used during training, with inference on Base model.
\vspace{8pt}

\noindent\textbf{TAD Visualization}:
Models cluster LQ image features, creating erroneous matching. 
Without reliable HQ-LQ image pairs in our target dataset, we use synthetic artifacts to demonstrate the LQ clustering on Celeb-ReID. 
\underline{\Cref{fig:lr_hr_cluster}} shows t-SNE plots of the LTCC pre-trained model on the Celeb-ReID dataset. 
TAD tries to make LQ and HQ features indistinguishable, eliminating LQ clusters (red dots), ideally overlapping with HQ features. 
This indicates robustness against LQ artifacts.
All three models - RLQ, CAL, and the Base Model - experience a significant drop in accuracy due to motion blur as shown in \underline{\Cref{fig:model_analysis} (left)}, specifically addressing it is \textit{left as future work}.
\vspace{8pt}

\noindent \textbf{Error Analysis} 
We cluster pose vectors using K-means with optimal performance at around 15 clusters as indicated \underline{\Cref{fig:model_analysis} (mid)}. We have two major spikes in accuracy, around the 15th and 25th clusters.
\underline{\Cref{fig:cal_gender_error2} (right)} shows a quantitative verification of gender awareness in RLQ compared to the Base Model and CAL. 
On their respective LTCC mispredictions, RLQ scores a 0.68 F-1 score on female classification (minority class), compared to CAL's 0.44 F-1 score and BM (0.61 F-1 score). 
\underline{\Cref{fig:cal_vs_rql}} instances where CAL misidentifies (a) gender, (b) pose (similar/flipped body), and (c) LQ images (motion blur matches motion blur), while RLQ correctly identifies.
When both models make mistake (d), RLQ correctly identifies gender and selects different poses.


%% file: sec/conlcusion.tex
\begin{table}[t!]
\centering
\renewcommand{\arraystretch}{1.1}
\setlength\tabcolsep{6pt}
\scalebox{0.9}{
\begin{tabular}{c|cc}
    \toprule
Model &R-1  & mAP \\   
\hline 
CAL &  51.7 & 8.9 \\ 	
Base Model & 58.8 & 14.4\\	
\rowcolor{LGray}
Base Model + CAP & 59.1		& 14.8\\
\hline 
\bottomrule
\hline 
\bottomrule
\end{tabular}
}
\vspace{-5pt}
\caption{\textbf{Celeb-ReID results}. TAD not applicable (already HQ).}
\label{tab:celeb_results}
\vspace{-7pt}
\end{table}

\subsection{Limitation}
\label{sec:limit}
We mainly focus on LQ images, deliberately disregarding fine-grained details that are available in HQ datasets.
Coarse Attribute Prediction (CAP) tries to extract maximum information from artifact-ridden inputs by approximating fine-grained attributes as coarse labels. 
Coarse labels can fall short compared to their fine-grained counterparts. 
On evaluating CAP on Celeb-ReID (\cref{tab:celeb_results}), CAP offers marginal improvements over the Base model.
We hypothesize the model has maximized its information gain from HQ RGB alone, leaving little scope for enhancement with coarse attributes. 

\noindent \textbf{Ethical Statement}
Our research targets real-world problems in ReID.
However, we recognize the potential risk for misuse in tracking and targeting individuals.
To safeguard against this, we will release some aspects of codes only via emails, for academic institutions, ensuring work's full potential is limited to academic research. 

\section{Conclusion}
In this work, we focus on the challenging problem of Clothes Changing Re-IDentification (CC-ReID) in real-world scenarios, where existing ReID models often struggle with low-quality artifacts. These artifacts, including pixelation, out-of-focus, and motion blur, introduce noise into fine-grained biometric attributes, complicating the learning of robust ReID features. Additionally, ReID models encounter difficulties in distinguishing between low-quality images, often leading to clustering and matching regardless of identity.
To overcome these challenges, we introduced a novel Robustness against Low-Quality (RLQ) framework. This framework leverages external auxiliary information through Coarse Attribute Prediction (CAP) to discretize noisy fine-grained attributes into more manageable representations. Simultaneously, the framework incorporates Task Agnostic Distillation (TAD) for robust internal feature representation by utilizing self-supervision and synthetic low-quality images.
We demonstrate the effectiveness of our approach on four real-world CC-ReID benchmarks.


%% file: Supp/arch.tex
\section{More on CAL Sensitivity problem}
\label{sec:CAL_sensativity}
~\Cref{tab:cal_sensativity} demonstrates the use of Coarse Attribute Prediction (CAP) for pose and gender in our Base Model and stand-alone CAL~\cite{gu2022clothes}.
Similar to Triplet loss and foreground augmentation, (shown in main submission), CAL is overly sensitive to foreign loss functions, leading to incompatibility. 
This strengthens our choice for a two-branch structure (Base Model) where we can learn attributes like pose and gender via a second branch (the BOT Branch) and communicate the information by sharing the backbone. 

\begin{table}[!h]
  \centering
  \begin{tabular}{@{}
  p{0.8cm}| 
  @{}P{0.92cm}@{}P{0.92cm}@{}| 
  @{}P{0.92cm}@{}P{0.92cm}@{}||
  @{}P{0.92cm}@{}P{0.92cm}@{}| 
  @{}P{0.92cm}@{}P{0.92cm}@{}}
    \toprule
    \multirow{3}{*}{Aux.} & 
    \multicolumn{4}{@{}c@{}||}{CAL} & \multicolumn{4}{@{}c@{}}{Base Model} \\ 
    \cline{2-9}
    & \multicolumn{2}{@{}c@{}|}{LTCC} & \multicolumn{2}{@{}c@{}||}{PRCC}  
    & \multicolumn{2}{c|}{LTCC} & \multicolumn{2}{c}{PRCC} \\
    \cline{2-9} 
    & R-1 & mAP & R-1 & mAP & R-1 & mAP & R-1 & mAP \\
    \hline 
    Vanilla &  40.1 & 18.0 & 55.2 & 55.8 & 41.1 & 20.3 & 58.0 & 59.4 \\ 
    \hline 
    +G & 34.9 & 17.0 & 45.6 & 45.5 & 43.5 & 21.7 & 58.8 & 59.8 \\ 
    +P & 34.2 & 15.9 & 43.5 & 41.5 & 42.5 & 20.2 & 59.4 & 60.6 \\ 
    +CAP & 36.0 & 16.8 & 43.7 & 43.5 & 43.4 & 20.7 & 60.2 & 60.3 \\ 
  \bottomrule
  \end{tabular}
  \caption{Contribution of each architectural component on vanilla CAL vs two-branch Base Model (CAP is $+$Gender$+$Pose)}
  \label{tab:cal_sensativity}
\end{table}

From the CAL's original paper, their ablation study shows extensive incompatibility of CAL with Triplet loss, the most commonly used ReID loss function (\cref{tab:cal_sensativity2}).

\begin{table*}[!t]
  \centering
  \setlength\tabcolsep{5pt}
  \begin{tabular}{@{}
  p{2cm}| 
  P{0.8cm}P{0.8cm}| 
  P{0.8cm}P{0.8cm}|
  P{0.8cm}P{0.8cm}| 
  P{0.8cm}P{0.8cm}|
  P{0.8cm}P{0.8cm}| 
  P{0.8cm}P{0.8cm}}
    \toprule
    \multirow{3}{*}{Method} & 
    \multicolumn{4}{c|}{LTCC} & \multicolumn{4}{c|}{PRCC} & 
    \multicolumn{4}{c}{CCVID~\cite{gu2022clothes}} \\ 
    \cline{2-13}
    & \multicolumn{2}{c|}{CC} & \multicolumn{2}{c|}{genral}  
    & \multicolumn{2}{c|}{CC} & \multicolumn{2}{c|}{SC} 
    & \multicolumn{2}{c|}{CC} & \multicolumn{2}{c}{general} \\ 
    \cline{2-13} 
    & R-1 & mAP & R-1 & mAP & R-1 & mAP & R-1 & mAP & R-1 & mAP & R-1 & mAP \\
    \hline 
    Vanilla &  40.1 & 18.0 & 74.2 & 40.8 & 55.2 & 55.8 & 100 & 99.8 & 81.7 &  79.6 & 82.6 & 81.3 \\ 
    + Triplet Loss & 34.7 {{(5.4$\downarrow$)}} & 16.6 {{(1.4$\downarrow$)}} & 71.8 {{(2.4$\downarrow$)}} & 37.5 {{(3.3$\downarrow$)}} & 48.6 {{(6.6$\downarrow$)}} & 49.7 {{(6.1$\downarrow$)}} & 100 & 99.8 & 
    81.1 
    {{(0.6$\downarrow$)}}
    & 77.0 
    {{(2.6$\downarrow$)}}
    & 81.5 
    {{(1.1$\downarrow$)}}
    & 78.1
    {{(3.2$\downarrow$)}}\\ 
  \bottomrule
  \end{tabular}
  \caption{Effect of Triplet loss on CAL}
  \label{tab:cal_sensativity2}
\end{table*}

\section{Base Model Details}
\label{sec:baseModel}

\begin{figure}[!h]
  \centering
  {\includegraphics[width=1\linewidth]{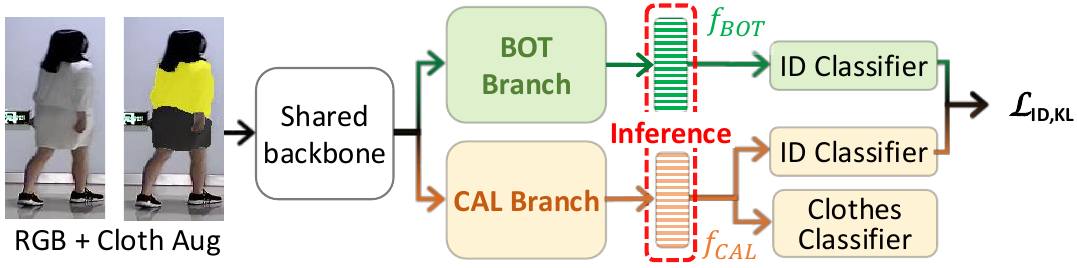}
  \caption{\textbf{Base Model}:  performance on `male' gender classification for all the datasets. The sharp drop in accuracy for LaST and DeepChange datasets indicates the effect of low-quality images on the off-the-shelf Face recognition model InsightFace~\cite{deng2019arcface}} 
  \label{fig:base_model}
  }
\end{figure}

\Cref{fig:base_model} shows our base model for learning clothes invariant features.  
CAL branch learns clothes invariant features, which helps the BOT branch via \textit{shared backbone}, and KL Divergence loss on identity classifier of both the branches $L_{ID,KL}$. 
BOT branches simply stabilize CAL adversarial learning (as described as "CAL sensitivity problem"). 
BOT Branch also learns clothes invariant features Clothes augmented data points, will are likely noisy. 
Such two-branch architectures have also been studied in existing works ~\cite{huang2021clothing, Hong_2021_CVPR, 9469545, 10.1007/978-981-99-7549-5_16, 10036012}. 
We adopt Huang \etal~\cite{huang2021clothing} method of ``Max-Avg Pooling'' (concat of global max and average pool) to produce $f_{CG}, f_{CA} \in \mathbb{R}^{4096}$ from both the branches. 
Concatenation of these features is used for inference. 
Identity logits (via Identity classifiers) $y^{ID}_{CG}, y^{ID}_{CA}$ predict the identity of the person during training only. 
We shall collectively refer to these two branches as \textbf{Base Model}.
Both these branches are briefly summarized below.

\noindent \textbf{CAL Branch:}
This branch focuses on clothing information (orange in the figure) and implements  CAL~\cite{gu2022clothes}, an RGB-only model with Bag-of-Tricks (BOT)~\cite{luo2019bag} framework (excluding Triplet and Center Loss). 
It uses an additional clothes classifier that predicts cloth labels ($y^{CL}_{CA}$). 
A clothes Adversarial Loss ($\mathcal{L}_{CAL}$) penalizes the clothes-relevant features for correct clothes predictions ($\mathcal{L}_{CL,CE}$), thereby generating cloth robust features. 

\noindent \textbf{BOT Branch:}
Standalone BOT Branch (green in the figure) is a simple ReID model (dilated strides in ResNet-50 last block, with Triplet Loss ($\mathcal{L}_{Trip}$)), similar to BOT.
This structure is commonly adopted by almost all ReID models due to its compatibility with different loss functions. In our work, we use it to disentangle pose and recognize gender (identity-related features).

\section{Clothes Augmentation (part of Base Model)}

\begin{figure}[!h]
\centering
{{\includegraphics[width=3cm]{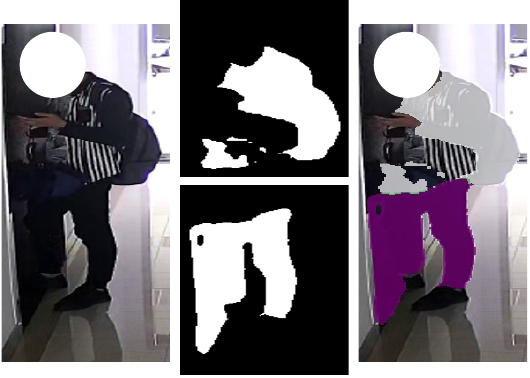}
\label{fig:clothes aug1} }}
\quad 
{{\includegraphics[width=3cm]{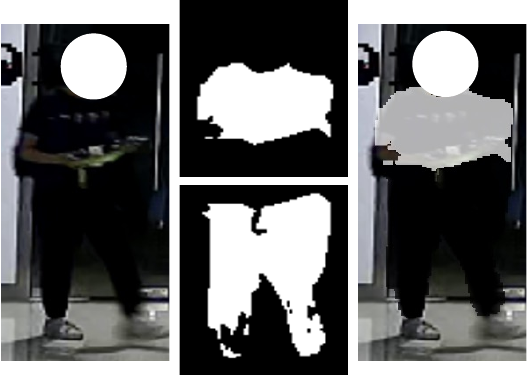}
\label{fig:clothes aug2} }}
\quad 
{{\includegraphics[width=3cm]{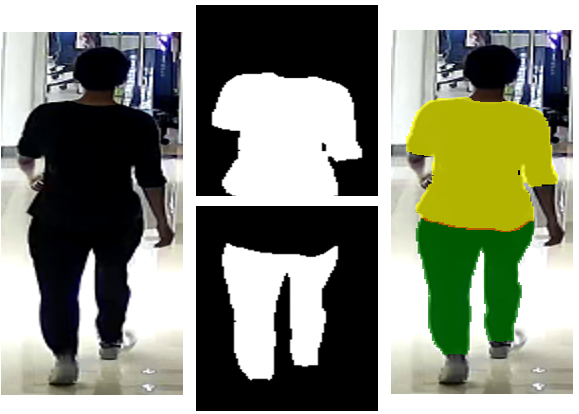}
\label{fig:clothes aug3} }} 
\quad  
{{\includegraphics[width=3cm]{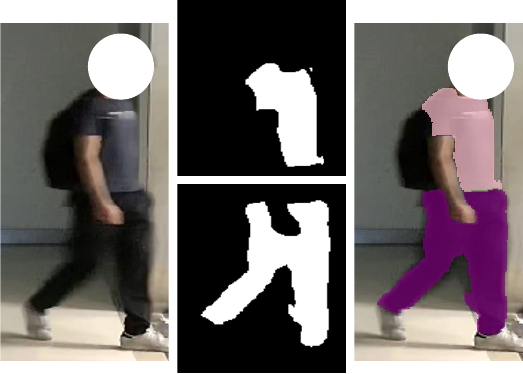}
\label{fig:clothes aug4} }}
\caption{\textbf{Clothing In-paint:} Upper-Lower Body Masks by SCHP~\cite{li2020self}, shown on LTCC Dataset. Random colors generate new clothing labels.} 
\label{fig:clothes}
\end{figure}
\label{sec:clothes_aug_vis}
A lot of ReID models use silhouettes-based body parsing for 
clothes-changing as an integral part of their framework. 
Contrary to this, 
our clothes-augmented samples as used only as additional augmentations and not part of our framework, because of the additional noise these silhouettes (fine-grained attributes) can bring to the model. 
Body parsing allows us to randomly change the upper and lower body colors as shown in \Cref{fig:clothes}. 
Noisy silhouette-based clothes augmentation does not affect the identity (\emph{BOT Branch}), gender (\emph{Gender Classifier}), or pose (\emph{POSE Branch}). Specifically for the \emph{CAL Branch}, these clothes have been given their separate clothing labels.

\section{Design Choices (More experiments)}
\label{sec:arch_choices}
\Cref{tab:backbone_split} shows the optimal split of the shared backbone between CAL and BOT branches occurs after the second block. 
\Cref{tab:pose_split} shows the split of Pose Branch, where we make the architectural choice of split after the first block based on the highest Top-1 accuracy.  \\

\begin{table}[!h]
\begin{minipage}{1\linewidth}
\centering
\begin{tabular}{p{2.3cm}|P{0.65cm}P{0.65cm}|P{0.65cm}P{0.65cm}}
\toprule
\multirow{2}{=}{Shared Backbone Split} 
& \multicolumn{2}{c|}{LTCC} 
& \multicolumn{2}{c}{PRCC} \\ \cline{2-5}
& R-1 & mAP & R-1 & mAP \\
\hline 
Disjoint & 40.1 & 19.4 & 57.9 & 59.3\\
After $1^{st}$ Block & 39.8 & 19.5 & 57.3 & 59.1\\ 
After $2^{nd}$ Block  & \textbf{41.1} & \textbf{20.3} & 58.0 & \textbf{59.4} \\
After $3^{rd}$ Block  & \textbf{41.1} & 19.0 & \textbf{58.8} & 58.6  \\
\hline 
\bottomrule
\end{tabular}
\caption{\textbf{Backbone Split}: Disjoint  means  separate backbones (Base Model) }
\label{tab:backbone_split}
\end{minipage}%
\\\hfill 
\begin{minipage}{1\linewidth}
\centering
\begin{tabular}
{
P{1.8cm}
|P{0.65cm}P{0.65cm}|P{0.65cm}P{0.65cm}}
\toprule
\multirow{2}{=}{Pose Branch \newline Split } &\multicolumn{2}{c|}{LTCC}&\multicolumn{2}{c}{PRCC} \\   
\cline{2-5} 
& R-1 & mAP & R-1 & mAP \\    
\hline 
$1^{st}$ Block & \textbf{46.4} &  21.5  &  \textbf{64.0} &  63.2 \\ 	
$2^{nd}$ Block & 44.4 & 21.5 & 62.6 & \textbf{64.2} \\	
$3^{rd}$ Block & 45.9 & \textbf{21.6} & 63.8 & 63.9 \\
\hline 
\bottomrule
\end{tabular}
\caption{\textbf{Pose Branch Split} on shared base in RLQ.}
\label{tab:pose_split}
\end{minipage}%
\end{table}

%% file: Supp/motivation.tex
\section{Effect of Low Quality Images on Soft Biometrics}
\label{sec:soft_biometrics}

\begin{figure}[!ht]
  \centering
  {\includegraphics[width=\linewidth]{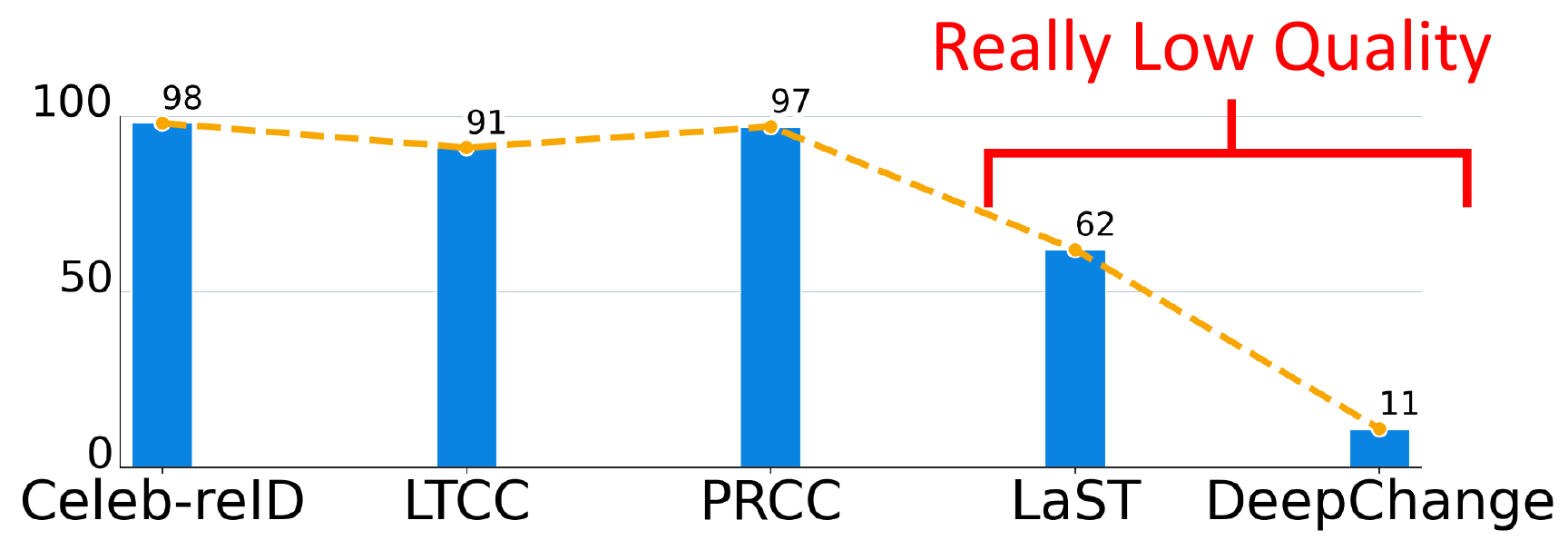}
  \caption{\textbf{Gender (soft biometrics}:  performance on `male' gender classification for all the datasets. The sharp drop in accuracy for LaST and DeepChange datasets indicates the effect of low-quality images on the off-the-shelf Face recognition model InsightFace~\cite{deng2019arcface}} 
  \label{fig:gender_fail}
  }
\end{figure}

Traditionally gender classification is used during inference time to filter out wrong genders for increasing ReID accuracy, commonly referred to as ``Soft Biometrics"~\cite{galiyawala2018person, lin2019improving}.  
\Cref{fig:gender_fail} shows the effect of low-quality images on the gender recognition capability of Face Recognition Model InsightFace~\cite{deng2019arcface}. This is one of the key reasons we opt for manual annotation of gender labels in the training set.

\begin{figure*}[!ht]
  \centering
  \includegraphics[width=\linewidth]{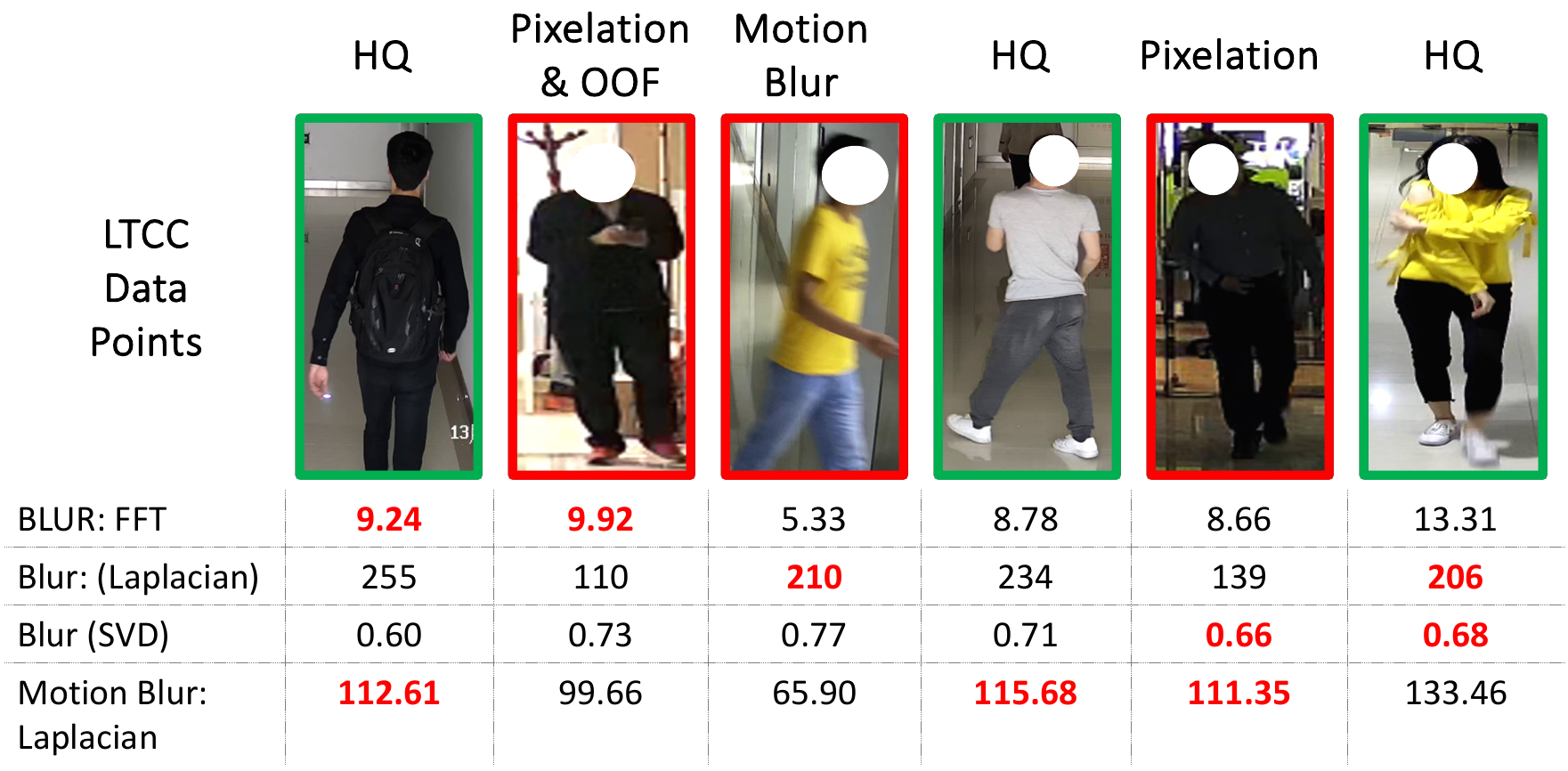}
  \caption{Edge detection scores for Fast Fourier Transform (FFT), Laplacian, Singular value decomposition (SVD), for various artifacts. 
  High-quality (HQ) images (shown in green) 
 should have a distinctively separate value compared to low-quality (LQ) images (shown in red), but there are similar conflicting values as shown in red.
}
  \label{fig:fail_detect}
\end{figure*}

\section{Analysis of high-quality and low-quality data (Failure of Traditional methods)}
\label{sec:sep_lr_hr}

Distinguishing between high-quality (HQ) and low-quality (LQ) data points, especially in the presence of artifacts like pixelation, motion blur, and out-of-focus blur, is a challenging task. 
While pixelation can be easily detected in images simply by looking at the spatial dimension ($\le 64 \times 64$), it’s not possible to identify pixelated images once they’ve been resized.
A common alternative for detecting these artifacts, thereby separating LQ images, is to use edge detectors such as Laplacian filters, Singular Value Decomposition (SVD), and Fast Fourier Transform (FFT), 
These tools score images, categorizing those above (or below) certain predefined thresholds as either blurry or high-quality. 
However, as shown in the \cref{fig:fail_detect}, these metrics often produce conflicting values (in red) for HQ (green) and LQ (red) data points, making it difficult to establish a clear boundary between the two.
Due to the lack of a consistent trend in these conflicting values, an external high-quality dataset, Celeb-ReID~\cite{huang2019celebrities}, is preferred for generating synthetic HQ-LQ image pairs and studying their relationships.

\begin{figure*}[!t]
  \centering
  \includegraphics[width=0.98\linewidth]{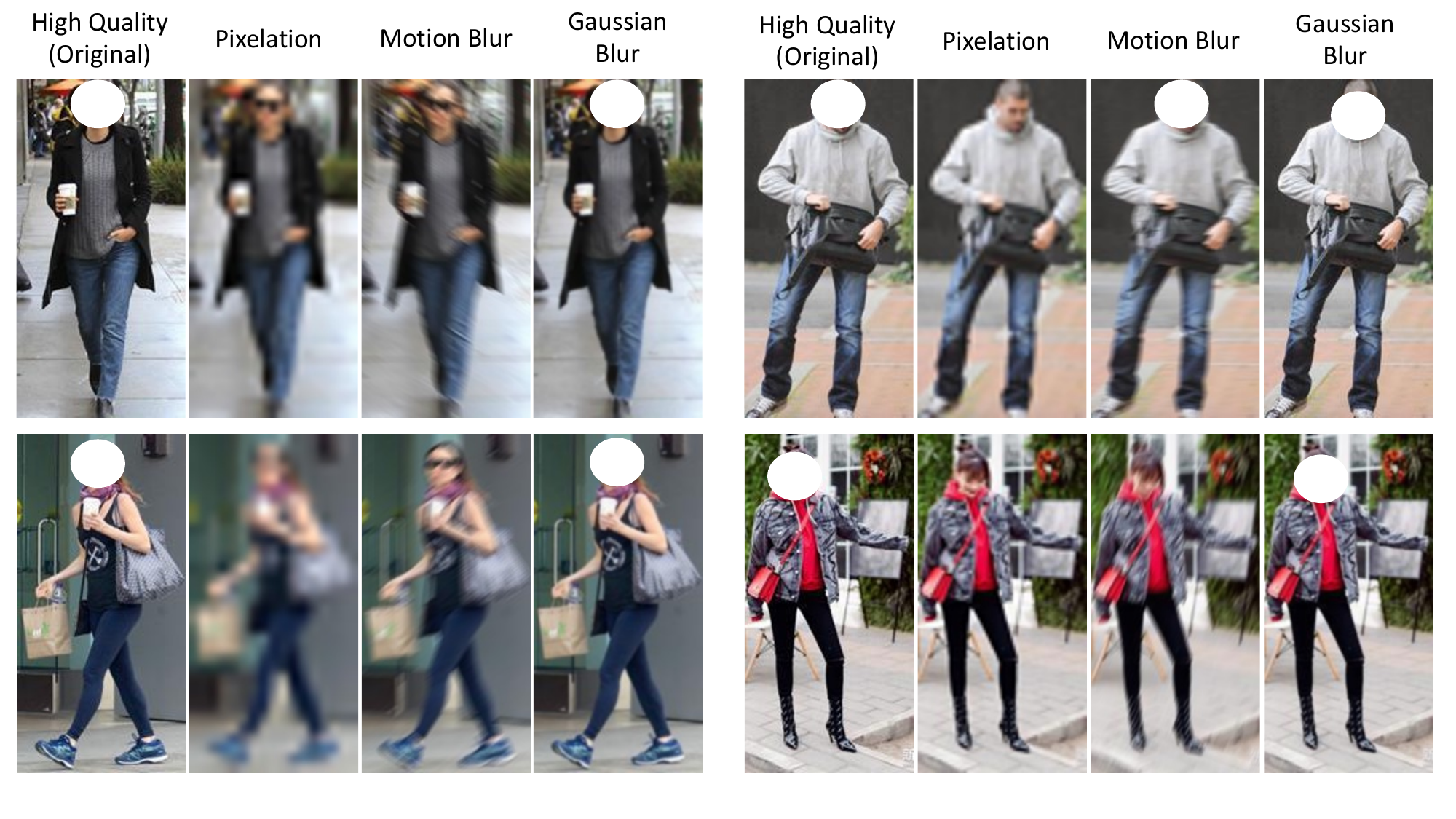}
  \caption{ Synthetic Low-Quality images, with images generated from Celeb-ReID. Gaussian blur simulates out-of-focus blur. }
  \label{fig:syntheric_lq}
\end{figure*}

%% file: Supp/training.tex
\section{Visualization of Synthetic Low Quality Images (Used in TBD) }
\label{sec:synthetic_lr}
We use Celeb-ReID~\cite{huang2019celebrities}, as a source of High-Quality images. 
We simulate real-world artifacts on it, namely pixelation, motion blur, and Gaussian blur (out-of-focus blur) as shown in ~\Cref{fig:syntheric_lq}. The method of generating these artifacts is explained in the \emph{under Implementation Details in main submission.}

\section{Additional Training Details }
\label{sec:add_training_details}
All branches are ImageNet pre-trained ResNet-50 blocks 
We use data augmentation like random horizontal flipping, erasing, and cropping along with clothes augmentation.  
Raw RGB images along with their synthetic clothing samples are normalized ($\mu=[0.485, 0.456, 0.406]$, $\sigma=[0.229, 0.224, 0.225]$).
We use Adam optimizer with warmup learning rate (lr=0.00035).
No hyperparameters were done, thus all loss terms were assigned a weight of 1.0.
Models are trained on with 4 positive samples per batch (for Triplet loss).
Gender labels can be manually generated ($0^{th}$ class in doubt), given the limited number of unique IDs in the training data, and errors in the genre recognition model on low-quality datasets. 
Pose vectors are computed on resized images, \ie resized poses. 
For TAD, the Teacher Base Model is pre-trained on the Celeb ReID dataset. 
All results are an average of two runs, with evaluation at every 10th epoch. We report the best Top-1 and mAP scores obtained throughout all the evaluations.

\section{Dataset Samples (RGB Examples) }
\label{sec:dataset_samples}
\begin{figure*}[!ht]
\centering
\subfloat[\centering LTCC \label{fig:ltcc_defects}]
{{\includegraphics[height=2.5cm]{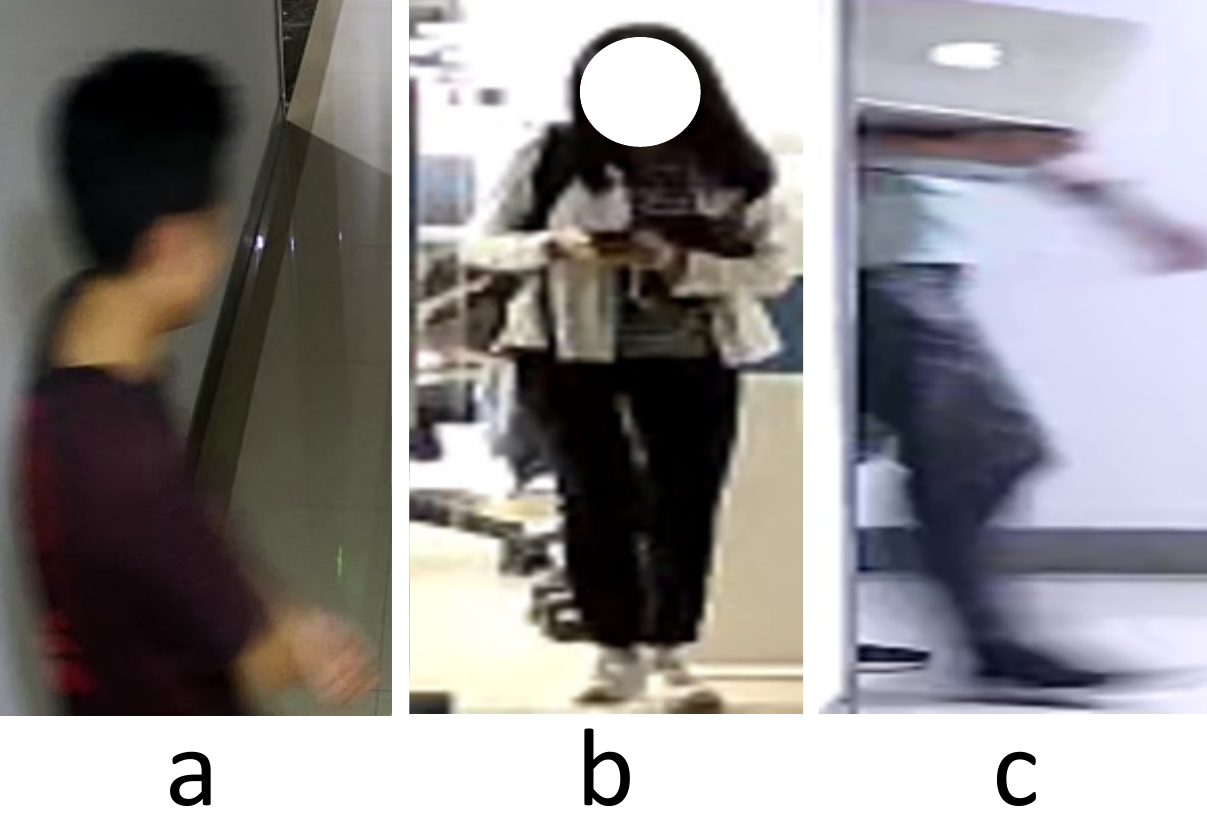}
}}
\hfill
\subfloat[\centering PRCC \label{fig:prcc_defects}]{{\includegraphics[height=2.5cm]{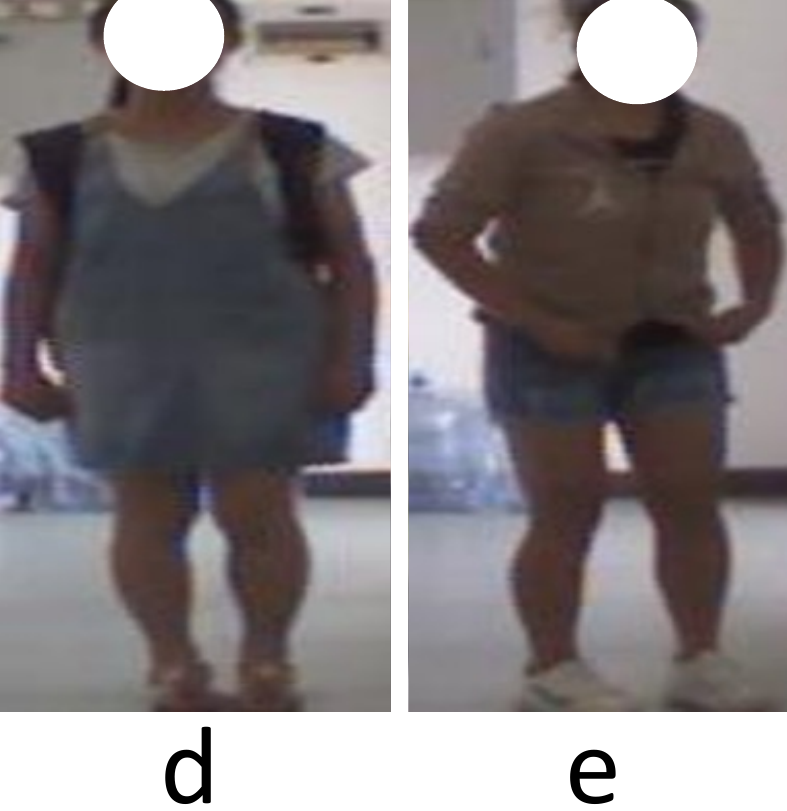}
}}
\hfill
\subfloat[\centering DeepChange \label{fig:deepchange_defects}]{{\includegraphics[height=2.5cm]{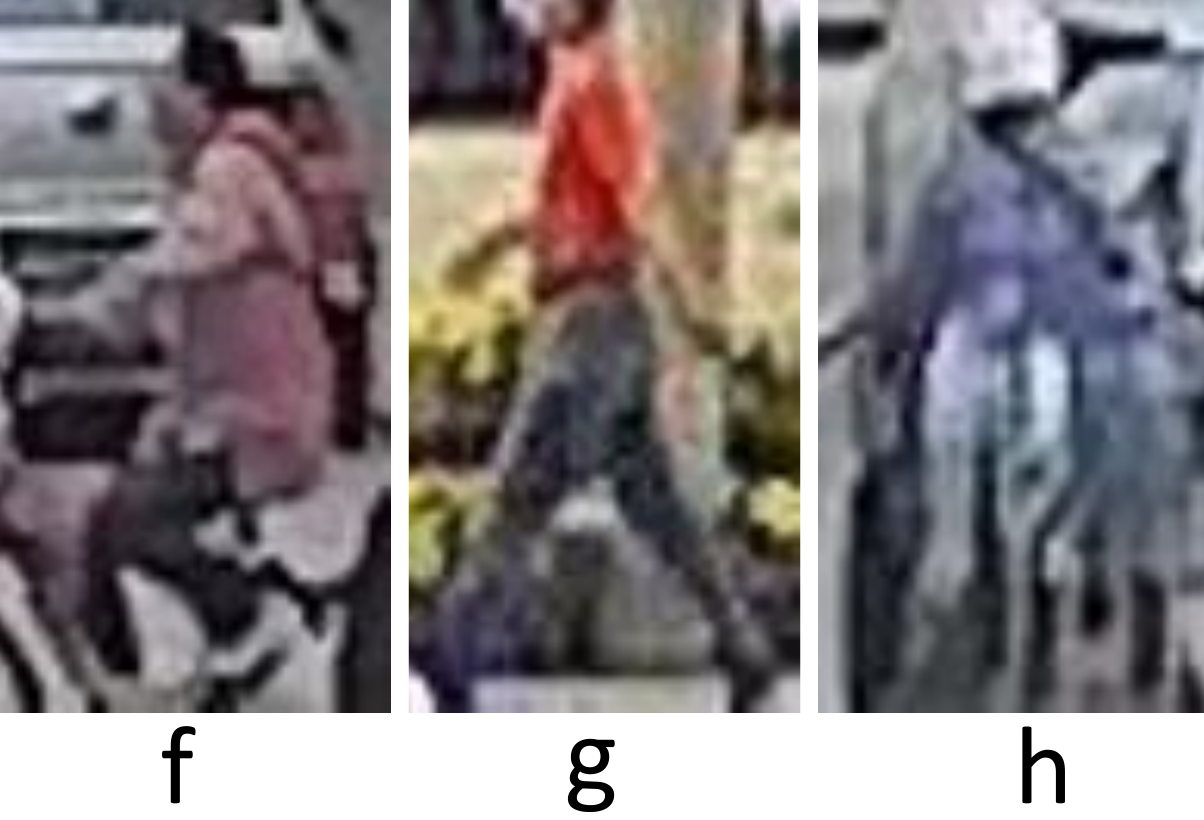}
}}
\hfill
\subfloat[\centering LaST \label{fig:last_defects}]{{\includegraphics[height=2.5cm]{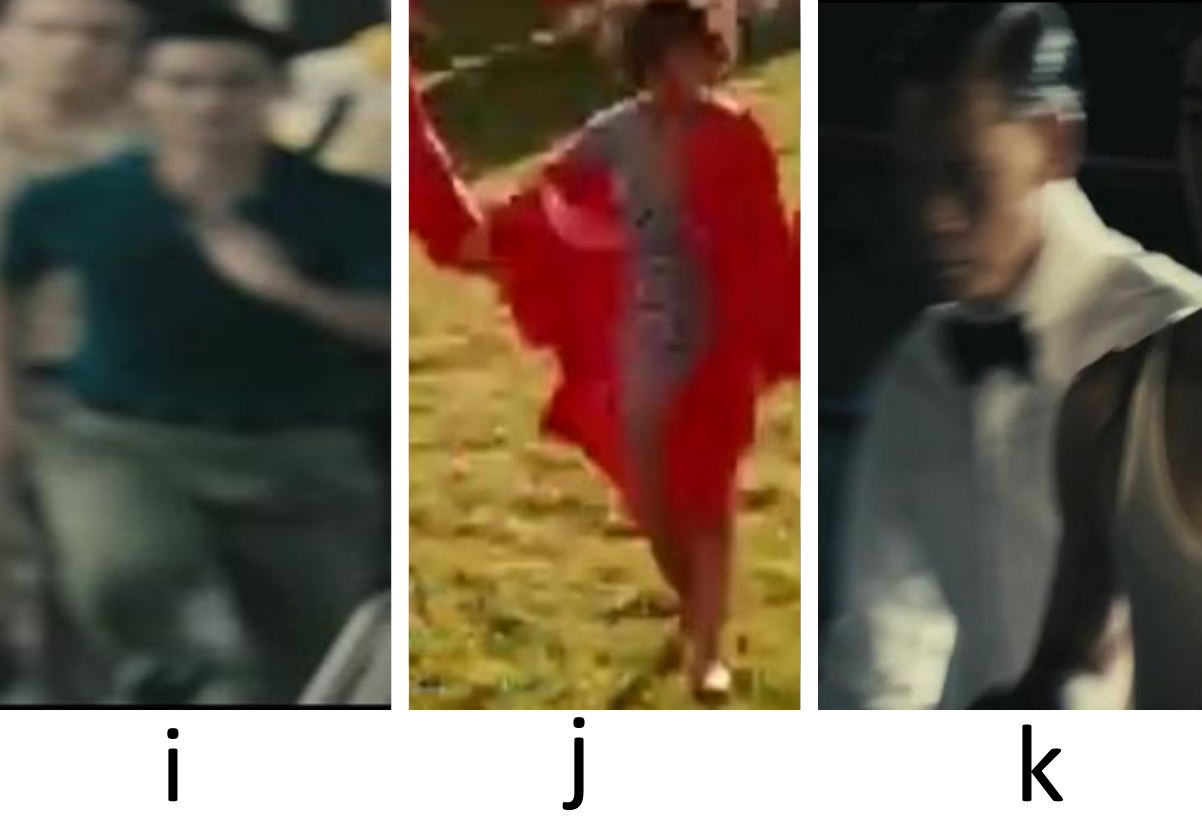}
}}
\caption{ 
\textbf{Dataset challenges:} 
 PRCC \& DeepChange have pixelation (b, f, g, h, j) and out-of-focus blur (a, d, e, k). LaST \& LTCC additionally have motion blur (c, i, k) 
 }%
\label{fig:Datasets_Defects}    
\end{figure*}

\textbf{LTCC}~\cite{qian2020long} has diverse poses from 12 indoor camera views, with noticeable motion, out-of-focus (OOF) blurring, and pixelation (\cref{fig:ltcc_defects}). \textbf{PRCC}~\cite{yang2019person} is relatively high-resolution and has some OOF and pixelation, featuring 3 indoor camera views (\cref{fig:prcc_defects}). \textbf{DeepChange}~\cite{xu2021deepchange} is captured across different weather conditions (12 months) in 17 camera views with massive pixelation and OOF (\cref{fig:deepchange_defects}). \textbf{LaST}~\cite{shu2021large} consists mostly of web/movie images with similar defects as LTCC (\cref{fig:last_defects}). 

%% file: Supp/vis.tex
\section{Visualization of Pose Clusters (RGB Examples) }
\label{sec:vis_pose_cluster}
\begin{figure*}[!ht]
\centering
\centering
\subfloat[\centering LTCC Pose Clusters]
{{\includegraphics[height=3.1cm]{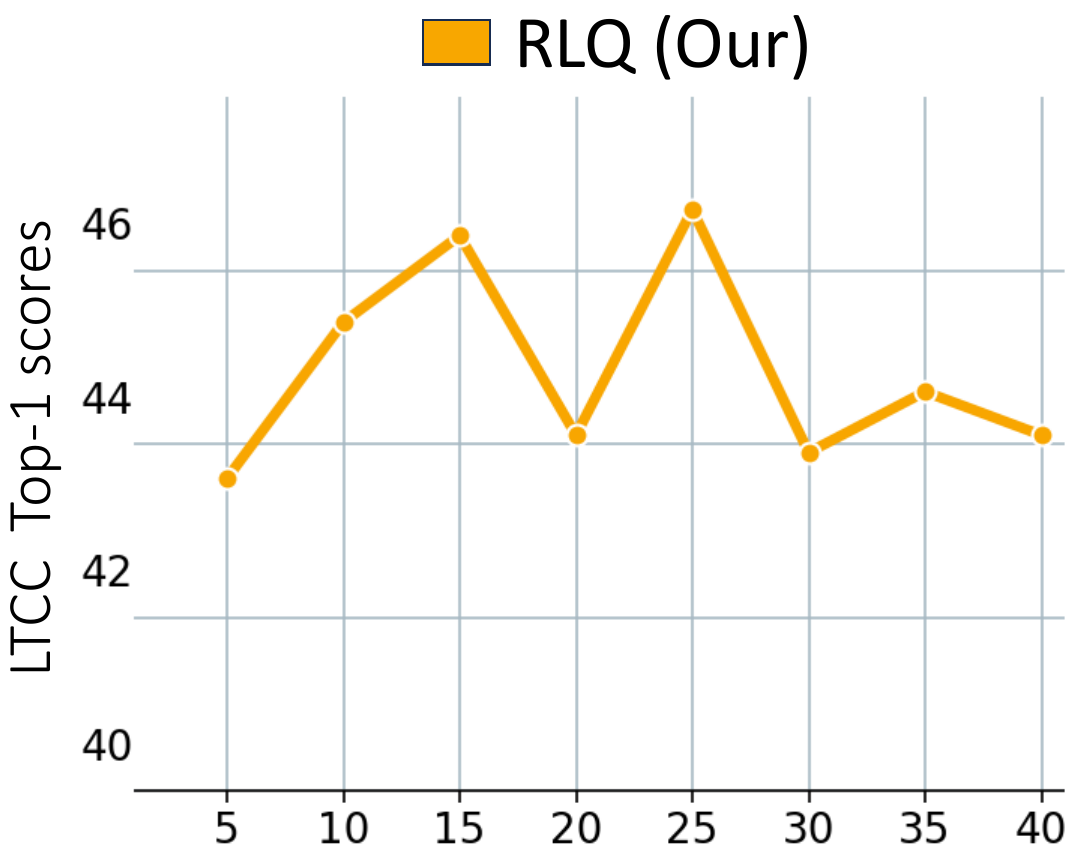}
\label{fig:ltcc_pose_clusters}
}}
\qquad 
\subfloat[\centering PRCC Pose Clusters]
{{\includegraphics[height=3.1cm]{Images/prcc_knn.pdf}
\label{fig:prcc_pose_clusters}
}}
\quad    
\caption{\textbf{Optimal K-means cluster size} of pose vectors on the LTCC and PRCC dataset.}
\label{fig:knn_size}
\end{figure*}

We obtain pose representation from off-the-shelf pose detector AlphaPose~\cite{fang2017rmpe}, which consists of key points (joints) and body lines (distance between each body joint). We cluster pose vectors using K-means with optimal performance at around 15 clusters as indicated ~\Cref{fig:knn_size}. We have two major spikes in accuracy, around the 15th and 25th clusters.

\begin{figure*}[!th]
\centering
\includegraphics[width=\linewidth]{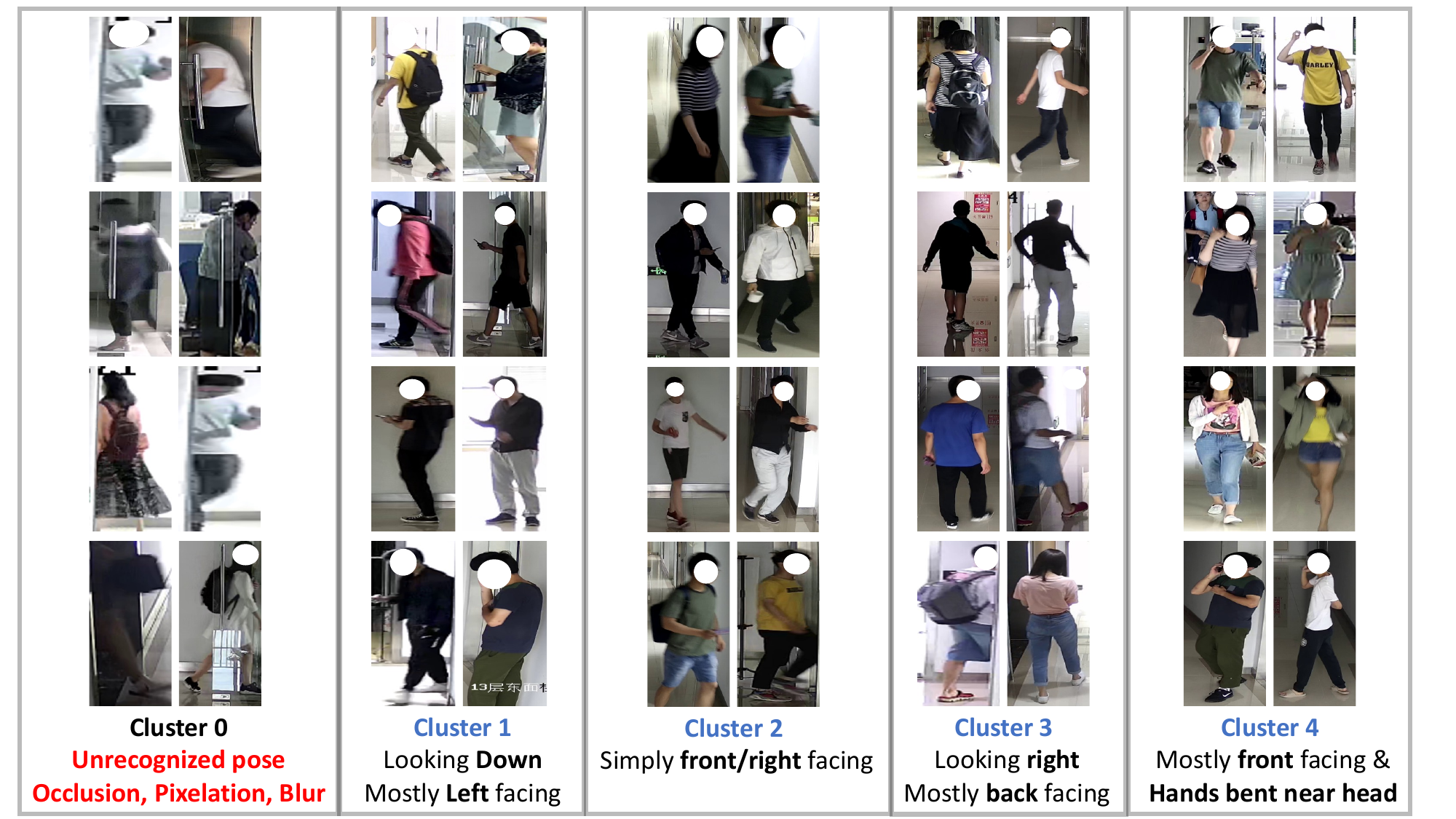}
\label{fig:pose_cl_0_3}
\caption{\textbf{Actual Pose Clusters} 
 Pose clustering on the LTCC dataset, with 15 clusters. Clusters are re-numbered to better fit.  Clusters from 0 - 4 are shown. }
\label{fig:pose_0_4}
\end{figure*}

\begin{figure*}[!tb]
\centering
{\includegraphics[width=\linewidth]{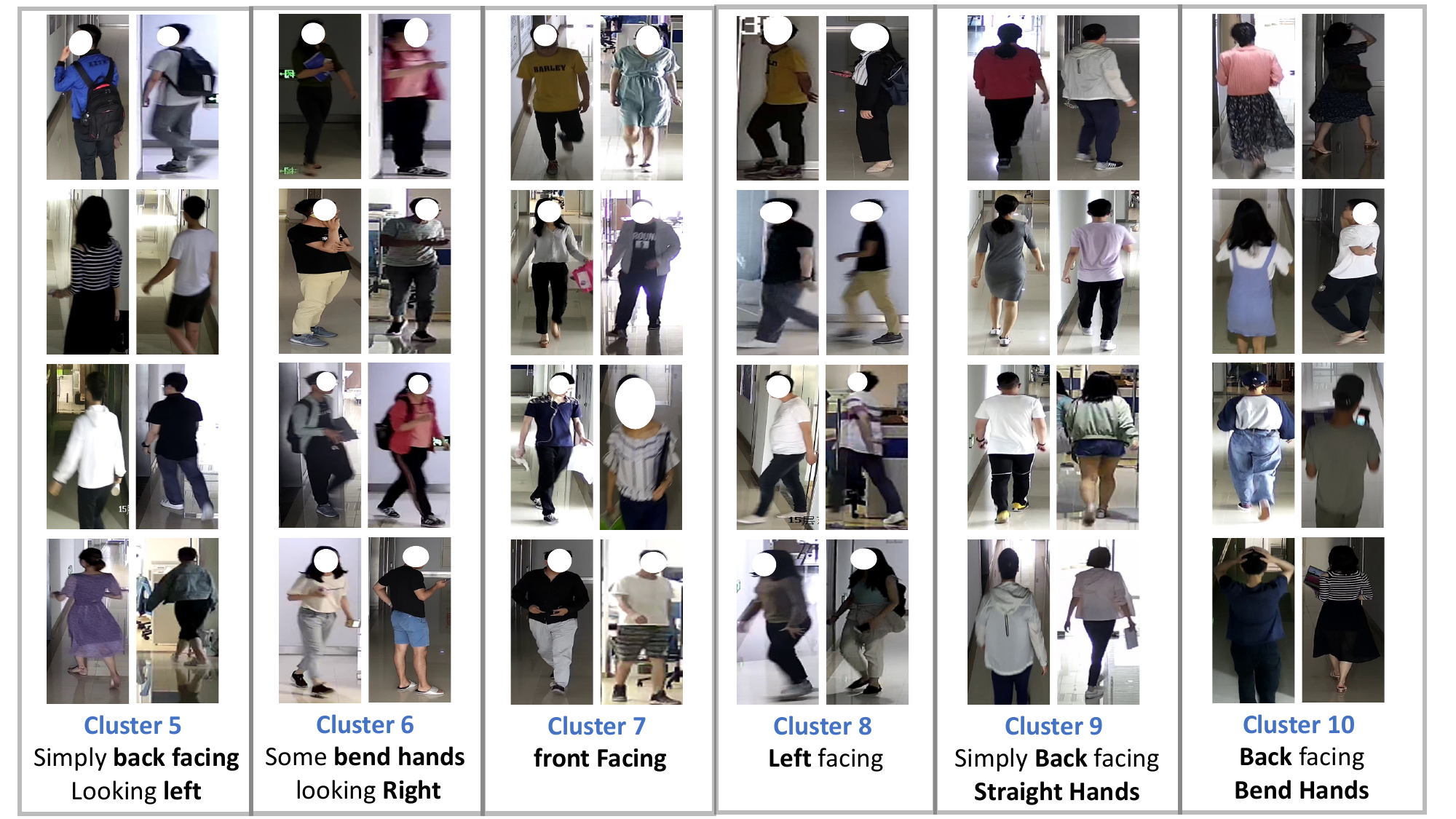}
\label{fig:pose_cl_8_11}}
\quad
{\includegraphics[width=\linewidth]{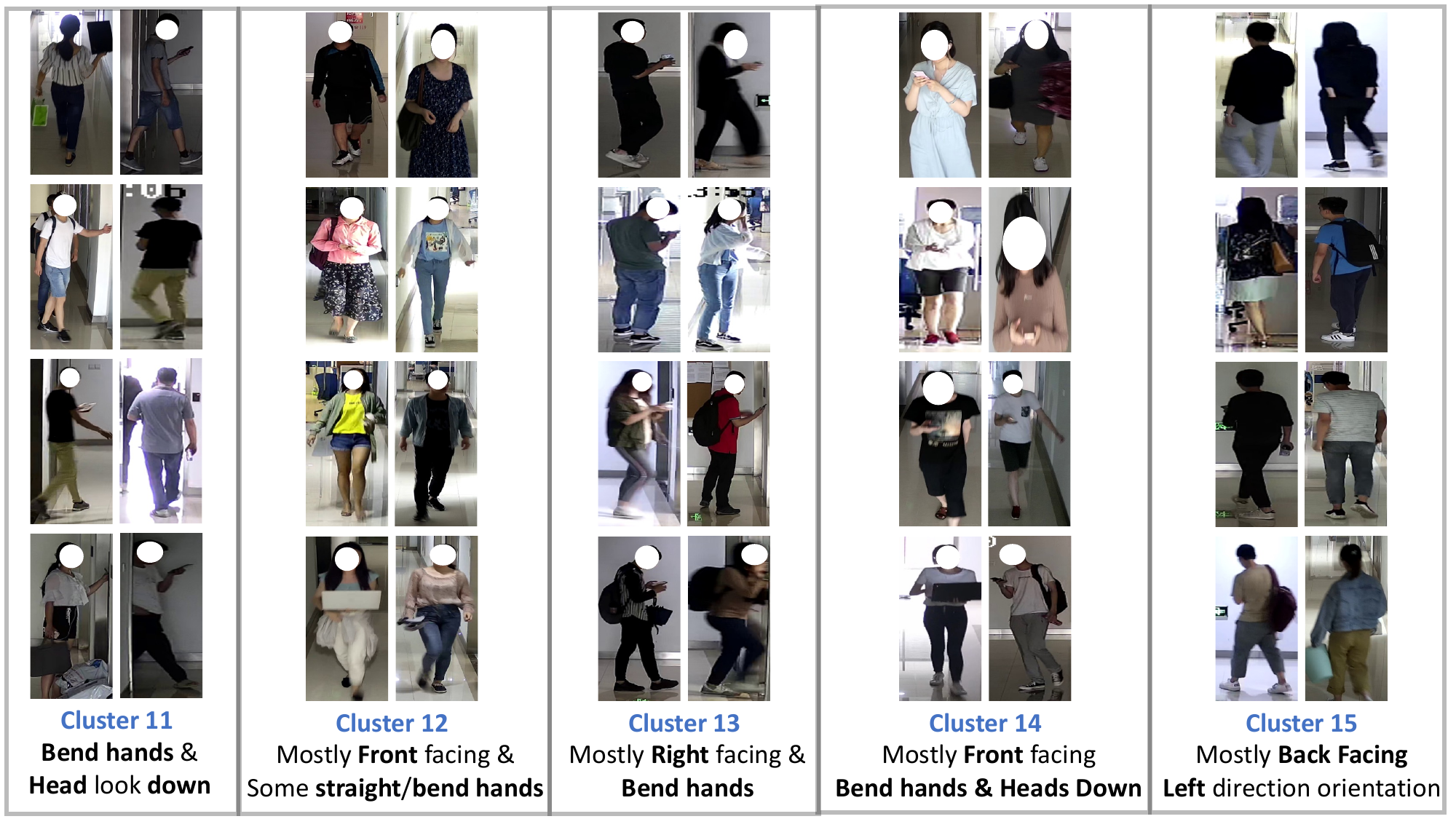}
\label{fig:pose_cl_12_15}}
\caption{ \textbf{Actual Pose Clusters:} Pose clustering on the LTCC dataset, with 15 clusters. Clusters are re-numbered to better fit.  Clusters from 5 - 15 are shown. }%
\label{fig:pose_5_15}
\end{figure*}

For better understanding, we have shown these pose clusters with their RGB images in ~\Cref{fig:pose_0_4} and ~\Cref{fig:pose_5_15}. 
We have also highlighted the commonalities across RGB images in their pose clusters. Pose Cluster 0 consists mainly of extreme outliers with most artifacts where the pose model couldn't detect any pose, thereby assigning the default 0th cluster.

